\documentclass[10pt,journal,compsoc]{IEEEtran}
\pdfoutput=1
\usepackage[pdftex]{graphicx}
\usepackage[cmex10]{amsmath}
\usepackage{amssymb}
\usepackage[ruled, norelsize]{algorithm2e}
\usepackage{bm}
\usepackage[caption=false,font=footnotesize]{subfig}
\usepackage{fancyhdr} 
\usepackage{algorithm2e}
\usepackage{booktabs}
\usepackage[export]{adjustbox}
\usepackage{balance}
\usepackage[table]{xcolor}

\ifCLASSOPTIONcompsoc
\usepackage[nocompress]{cite}
\else
\usepackage{cite}
\fi
\newcommand{\removelatexerror}{\let\@latex@error\@gobble}
\makeatother

\allowdisplaybreaks
\newcommand{\overbar}[1]{\mkern 3mu\overline{\mkern-3mu#1\mkern-3mu}\mkern 3mu}
\begin{document}
	
\title{Multistage Curvilinear Coordinate Transform Based Document Image Dewarping using a Novel Quality Estimator}

\author{Tanmoy Dasgupta,~\IEEEmembership{Member,~IEEE,}
	Nibaran Das,~\IEEEmembership{Member,~IEEE,}
	and Mita Nasipuri,~\IEEEmembership{Senior~Member,~IEEE}
	\IEEEcompsocitemizethanks{
		\IEEEcompsocthanksitem T. Dasgupta is with the Department of Electrical Engineering, Techno India University, West Bengal, India.\protect\\
		E-mail: tdg@ieee.org
		\IEEEcompsocthanksitem N. Das and M. Nasipuri are with the Department of Computer Science and Engineering, Jadavpur University, India.\hfil\break
		Emails: nibaran.das@jadavpuruniversity.in, mitanasipuri@gmail.com
	}
	\thanks{\null}}

\markboth{\null}%
{Dasgupta \MakeLowercase{\textit{et al.}}: Curvilinear Coordinate Transformation Based Document Image Dewarping Technique}

\IEEEtitleabstractindextext{	
\begin{abstract}
The present work demonstrates a fast and improved technique for dewarping nonlinearly warped document images. The images are first dewarped at the page-level by estimating optimum inverse projections using curvilinear homography. The quality of the process is then estimated by evaluating a set of metrics related to the characteristics of the text lines and rectilinear objects for measuring parallelism, orthogonality, etc. These are designed specifically to estimate the quality of the dewarping process without the need of any ground truth. If the quality is estimated to be unsatisfactory, the page-level dewarping process is repeated with finer approximations. This is followed by a line-level dewarping process that makes granular corrections to the warps in individual text-lines. The methodology has been tested on the CBDAR 2007 / IUPR 2011 document image dewarping dataset and is seen to yield the best OCR accuracy in the shortest amount of time, till date. The usefulness of the methodology has also been evaluated on the DocUNet 2018 dataset with some minor tweaks, and is seen to produce comparable results.
\end{abstract}

}

\maketitle
\IEEEdisplaynontitleabstractindextext
\IEEEpeerreviewmaketitle

\IEEEraisesectionheading{\section{Introduction}}
\IEEEPARstart{C}{apturing} document images with handheld devices, such as cameras, mobile phones, etc. often introduce warping in the acquired images, along with a plethora of other issues. Even if it's somehow managed to get all the preprocessing steps, such as contrast enhancement, binarization, etc. to yield perfect results, the warped texts in such document images still pose a severe challenge to standard OCR techniques, resulting in overall poor digitization of the documents. 

Various algorithms exist for correcting a camera-captured document image containing perspective defects. But the results often introduce unnecessary skews in the text characters. Standard OCR techniques often rely heavily upon the assumption that the baselines of the adjacent text-lines are parallel as far as practicable and that the text characters have minimal skews in them. This necessitates the use of dewarping algorithms that can take care of individual character-level warps after performing an overall page-level dewarping. 

There are several fast and effective techniques available for correctly dewarping linearly warped document images \cite{wu2002document, lu2006document, Mollah_Skew_correction}. However, these methods are not suitable for dewarping non-linearly warped document images \cite{shafait2007document}. To address the inherent complexity of non-linearly warped document images, a global optimization based dewarping technique was proposed by Ezaki \textit{et. al.} \cite{ezaki2005dewarping}. There, nonlinear warps were corrected by minimizing an objective function that strives to convert the warped lines to parallel ones. With a similar objective, curled text-line detection based techniques were demonstrated in \cite{ulges2005document, kakumanu2006document}.  These methods are, however, less robust that the boundary based model-fitting techniques which were introduced in \cite{Wu_Model, wu2008model, masalovitch2007usage, HePan, huggett2013method, bolelli2017indexing}. Gatos \textit{et. al.} \cite{Gatos} implemented a generalized approach that involved segmentation of the text-lines to determine the warps in the document images. This method performs better than the older ones in many use-cases. Generalizations, that involved representing the warped text-lines as a texture delimited by two smooth curved lines on top and bottom were described in \cite{CHETHAN2010330, Stamatopoulos,5585760, kwon2016method}. These methods involve a coarse but fast dewarping of the whole document image based on these two curves followed by finer corrections based on word detection. However, this technique heavily relies on the assumption that the processed image mostly contains text and not diagrams or pictures. Bukhari \textit{et. al.} \cite{Bukhari} demonstrated an active contour based warp correction technique for detection of curled text-lines. Although it produced better results in many cases, being a heavily iterative process, this methodology is considerably slower than many of the previous ones. A generalization of such model-based techniques was developed in \cite{Meng_Metric}, that uses a generalized cylindrical surface model to describe non-linearly warped pages. Depth information (if available) can also be used to dewarp a document image \cite{Wu_2015}. Such methodologies work for particular use cases and yield unsatisfactory results when applied to arbitrarily warped document images. Slightly more robust algorithms based on text-line detection \cite{Frinken_Word_Spotting, Koo_text-line} were introduced in \cite{KIM20153600, Kil} for handling complex page layouts. These methods iteratively check the alignment of the text-lines and try to improve it by dewarping them. Although such techniques can outperform the older ones, they are inherently slow and reliant upon the availability of bounding boxes and discernible page-boundaries or page limits. Koo \textit{et. al.} \cite{koo2009composition} introduced a technique that can estimate and rectify warps by capturing the document image from multiple viewpoints. This was further improved in \cite{You_Multiview} by considering a ridge-aware 3d model for describing the warped documents. A similar, but slightly more constrained technique was earlier developed in \cite{bukhari2009ridges}. Similar techniques based on stereo-vision were developed earlier in \cite{ulges2004document, cutter2012capture}.

A recent trend is the application convolutional neural networks for dewarping specifically warped documents. For example, \cite{Ma_DocUNet, Das_2019_ICCV, 2019arXiv190909470L} depict very effective ways of dewarping images of previously folded documents.  Most deep-learning based methods in this domain, however, suffer from the fact that number of available sample dataset is quite small \cite{shafait2007document, IUPR_dataset}. There is, in fact, a strong need in research towards developing synthetically warped document images \cite{6628669, Kieu_2013, Garai_Synthetically}.

It is quite evident from the above discussion that most methods are designed with specific goals in mind. Some of them produce better accuracy at the cost of speed. However, it is often found that camera captured document images contain parts of other pages that are supposed to be beyond the region of interest (ROI). Moreover, most of the techniques rely heavily on assumptions like structural uniformity of the warps, inherent parallelism of the text-lines, etc. These assumptions might not always be true. Also, many dewarping techniques introduce \emph{skews} in the characters as a side-effect of the dewarping process.

\begin{figure*}
	\centering
	\includegraphics[width=0.8\linewidth]{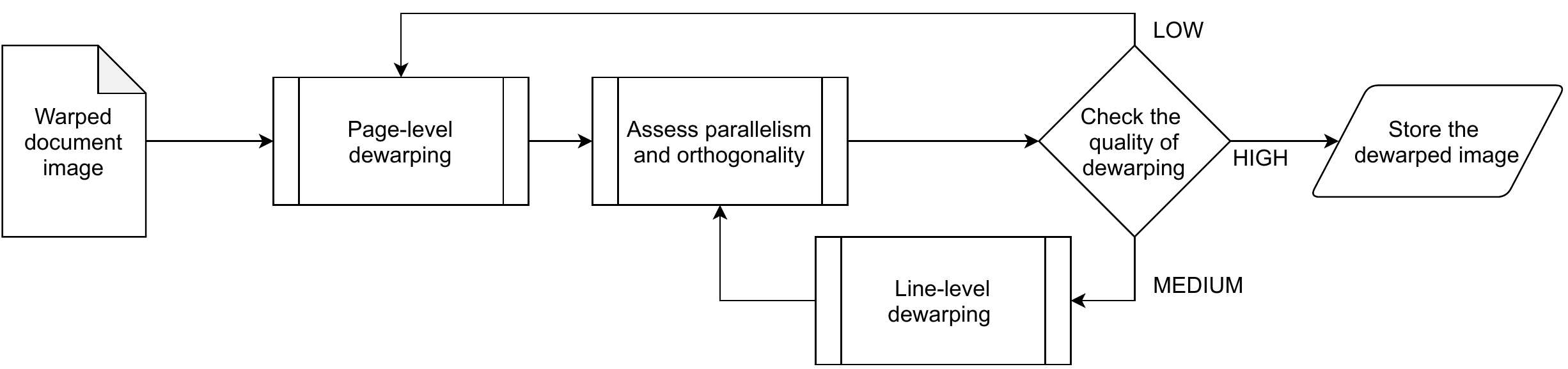}
	\caption{Representative block diagram of the dewarping process. The warped image is first dewarped at the page-level. Then quality of the result is assessed and if deemed necessary, the page-level dewarping is repeated. Otherwise, if required, line-level dewarping is applied on the processed image.}
	\label{fig:process}
\end{figure*}

The present technique demonstrates a multistage approach for dewarping non-linearly warped document image. First, a proper estimation of the warps in the document images is made through a piecewise linear approximation based generalization of the law of anharmonic ratio under homography. The underlying assumption is that, either the page boundaries \cite{Guo_Outline}, or at least, two lines of printed text would be clearly visible. If the page boundaries are unambiguously discernable, then they are used to make an overall estimate for the curvilinear homography. Otherwise, the underlying homography is estimated by analyzing text-lines that are supposed to be parallel in the original unwarped printed document. The estimation is then extrapolated to assess the page level warps. Once such an estimate is available, the whole document image is divided into small sections by generating a perspective transformation grid. This is followed by a page-level dewarping by generating optimum inverse projections for each block. The quality of this process is then assessed by calculating five metrics related to the characteristics of the text-lines and rectilinear objects for measuring the parallelism, orthogonality, etc. in the dewarped images. This is done irrespective of the availability of any ground truth. Based on the calculated metrics, if the result of page-level dewarping after a single iteration is found to be unsatisfactory, the process is repeated with finer adjustments. If it is realized that further page-level dewarping would only produce diminishing improvements, then a line-level dewarping akin to \cite{Roy_Warp, Adhikari_Unwarping, Garai_dewarp_bangla}, is applied that dewarps individual text-lines. Such considerations make the present methodology the fastest among all  presently available techniques. The whole process is depicted in fig. \ref{fig:process}. The programmes were tested in CBDAR 2007 / IUPR 2011 document image dewarping dataset \cite{shafait2007document, IUPR_dataset} and the performance of the methodology is evaluated on the basis of the performance of an OCR engine and the dewarping evaluation measure \cite{6320850} on the end result. After establishing the effectiveness of the presented methodology, the technique has also been tested on the DocUNet 2018 \cite{Ma_DocUNet} dataset containing 130 synthetically warped document images containing various kinds of non-uniform warps. The results are found to be quite promising.

The following are the specific contributions of the present work.
\begin{enumerate}
	\item A multistage dewarping algorithm is designed to perform coarse and fine adjustments as needed.
	\item A quick and accurate methodology is devised to easily extract the regions of interest (ROI) from a document image. This saves a lot of unnecessary work for the dewarping algorithms. 
	\item A mechanism has been developed for assessing the quality of the dewarped images without the need of ground-truth images. This has been used to shape the internal decision-making process. 
	\item The decision-making process is designed in a way that quickly determines the correct courses of action -- eliminating any step that might provide negligible improvements with large time-penalty.
	\item The results are analyzed thoroughly with the CBDAR 2007 / IUPR 2011 dataset to provide a detailed insight on the correlation between the different document layouts and types of warps with the performance of the methodology.  
	\item The effectiveness of the algorithms are further validated by testing them on the DocUNet 2018 dataset. This dataset is especially challenging, since, many of the assumptions related to inherent smoothness of the warps do not apply here. The dewarping results are found to be reasonably good on this dataset also.
\end{enumerate} 

The rest of the paper is organized as follows. Section 2 provides a detailed description of the present methodology. ROI selection, performance metrics and experimental results are discussed in Section 3. Section 4 concludes the discussion with remarks on future possibilities. 

\section{Present Methodology}
The dewarping process developed in this work is implemented in four stages. First, the law of anharmonic ratio under linear projections is analyzed. Then, a suitable generalization of the methodology for non-linear perspective transformation is developed, which is used on the document images to generate a nonlinear perspective grid depicting the curvilinear warps in the distorted images. A suitable optimization technique is then employed to quickly estimate the optimum inverse projections. This is done based on the calculation of certain parameters that can represent the distortions due to curvilinear warps. Based on the estimated inverse projections, page-level dewarping is applied on the images. Finer adjustments are then made by performing line-level dewarping that minimizes the skews in the dewarped text-lines. 
\subsection{Law of anharmonic ratio under homography for images of linearly warped rectangular pages}
\begin{figure}
	\centering
	\includegraphics[width=0.5\linewidth]{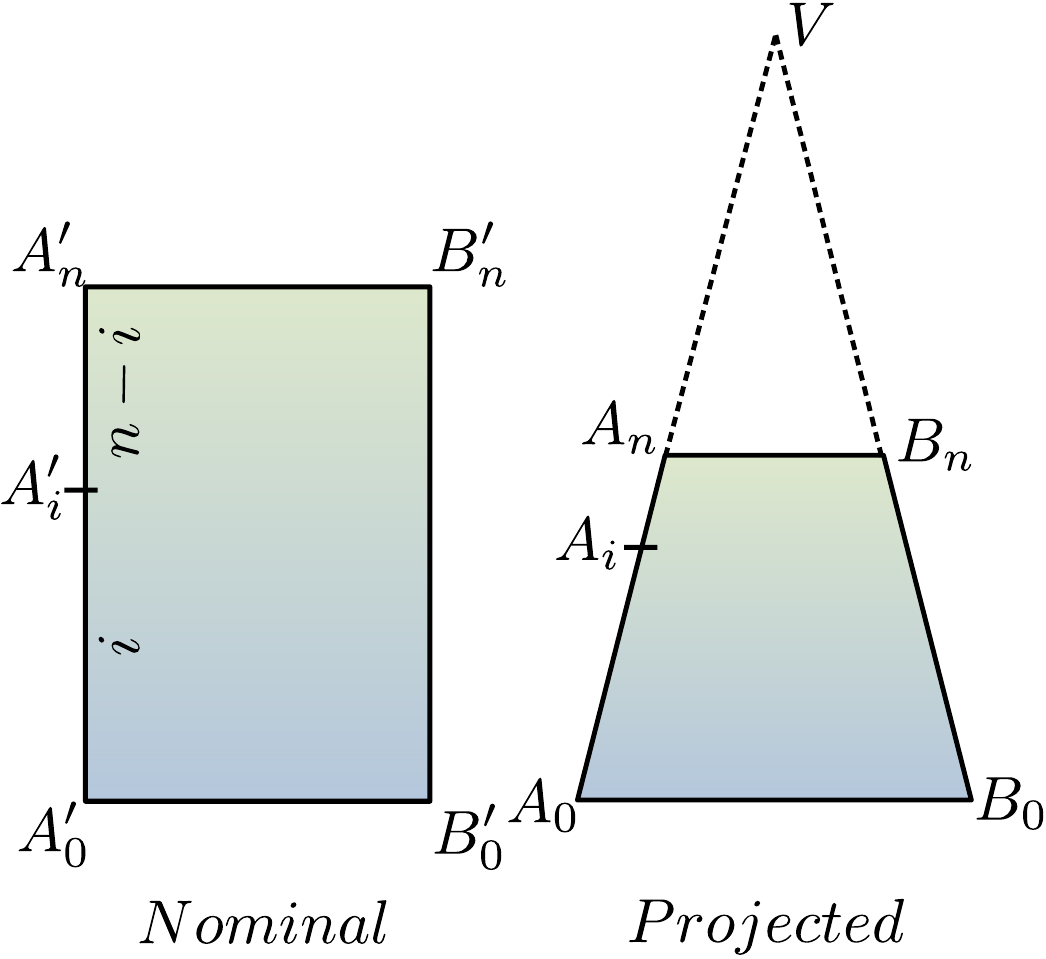}
	\caption{A rectangular page and its perspective transformation}
	\label{fig:homography}
\end{figure}

The proposed process of page-level dewarping is based on a generalization of linear homography for curvilinear warps. A linear homography would simply be a perspective transformation. For example, a rectangular page and an instance of it undergoing perspective transformation is depicted in fig. \ref{fig:homography}. 
In the real world, without any kind of perspective transformation, the long edges are denoted by $A_0'A_n'$ and $ B_0'B_n' $. Under a perspective transformation the edges become $A_0A_n$ and $ B_0B_n $. Let $ A_i' $ be a point on the edge $A_0'A_n'$ such that $ A_i' $ divides the line segment $A_0'A_n'$ in a $ i:(n-i) $ ratio. Thus, $ \dfrac{\overbar {A_0'A_i'}}{\overbar{A_0'A_n'}} = \dfrac{i}{n} $. In the projected plane, $ A_i' $ becomes $ A_i $. Now, consider that the extended versions of the lines $A_0A_n$ and $ B_0B_n $ in the projected plane meet at the \textit{vanishing point} $ V $. Now, using the theorem of anharmonic ratio, it can be written that,
\begin{align*}
&\dfrac{\overbar{A_0A_n}\times \overbar{A_iV}}{\overbar{A_iA_n}\times \overbar{A_0V}} = \dfrac{\overbar{A_0'A_n'}}{\overbar{A_i'A_n'}} = \dfrac{\overbar{A_0'A_n'}}{\overbar{A_0A_n} - \overbar{A_0'A_i'}} = \dfrac{n}{n-i}.\\
\therefore & \dfrac{\overbar{A_iV}}{\overbar{A_iA_n}} = \left( \dfrac{n}{n-i} \right)  \dfrac{ \overbar{A_0V}}{\overbar{A_0A_n}} = k_0\left( \dfrac{n}{n-i} \right), \quad \left[k_0 = \dfrac{ \overbar{A_0V}}{\overbar{A_0A_n}}\right].\\
\therefore &\dfrac{\overbar{A_iV}}{\overbar{A_iA_n}} - 1 = \dfrac{k_0n-n+i}{n-i},\\
&\textrm{and thus, } \overbar{A_iA_n} = \dfrac{n-i}{k_0n-n+i} \overbar{A_nV}.\\
&\textrm{This yields, }\dfrac{\overbar{A_0A_n} - \overbar{A_iA_n}}{\overbar{A_0A_n}} = 1 - \dfrac{n-i}{k_0n-n+i} \left( \dfrac{\overbar{A_nV}}{\overbar{A_0A_n}} \right). \\
\therefore & \dfrac{\overbar{A_0A_i}}{\overbar{A_0A_n}} = 1 - \dfrac{(n-i)\overbar{A_nV}}{n\overbar{A_0V} - (n-i)\overbar{A_0A_n}}.
\end{align*}
Thus, the point $ A_i' $ on the line segment $ \overline{A_0'A_n'} $ that actually divides  $ \overline{A_0'A_n'}$ in a way such that $ \dfrac{\overbar {A_0'A_i'}}{\overbar{A_0'A_n'}} = \dfrac{i}{n} $, is mapped to a point $ A_i $ in the projected plane such that, 
\begin{equation}
\label{eq:cross-ratio}
\dfrac{\overbar{A_0A_i}}{\overbar{A_0A_n}} = 1 - \dfrac{(n-i)\overbar{A_nV}}{n\overbar{A_0V} - (n-i)\overbar{A_0A_n}}.
\end{equation}
This is true for any projected line on the page that passes through the vanishing point. Also, the apparent symmetry in the projection as depicted in fig. \ref{fig:homography} is just for the purpose of keeping the diagram simple. It is not at all necessary for the theorem.  

If the vanishing point is closer to the edge $ A_0B_0 $ instead, then the equivalent ratio can be obtained by swapping $ A_0 $ and $ A_n $ in \eqref{eq:cross-ratio}. Thus,
\begin{equation}
\label{eq:cross-ratio2}
\dfrac{\overbar{A_iA_n}}{\overbar{A_0A_n}} = 1 - \dfrac{(n-i)\overbar{A_0V}}{n\overbar{A_nV} - (n-i)\overbar{A_0A_n}}.
\end{equation}
Let us assume that there is a line segment on the projected page with endpoints $ (x_0, y_0) $ and $ (x_n, y_n) $ that vanishes at $ V $. The set of points $ (x_i, y_i) $ for $ i=1, 2, \cdots, n-1 $, that actually split the line segment into $ n $ number of equal parts in the real world, can be determined as,
\begin{equation}
(x_i,\,\, y_i) = \left( x_0 + p_i(x_n - x_0),\,\, y_0 + p_i(y_n - y_0) \right) ,
\end{equation}
$ \textrm{for } i=1, 2, \cdots, n-1, $
where, 
\begin{equation}
p_i = 
\begin{cases}
1 - \dfrac{(n-i)\overbar{A_nV}}{n\overbar{A_0V} - (n-i)\overbar{A_0A_n}}, & \textrm{if } V \textrm{ is closer to } A_n\\
\dfrac{(n-i)\overbar{A_0V}}{n\overbar{A_nV} - (n-i)\overbar{A_0A_n}}, & \textrm{if } V \textrm{ is closer to } A_0\\
\end{cases}.
\end{equation}
\subsection{Generalized homography under curvilinear warps}
In situations, where the page is nonlinearly warped, homographies can be generalized through piecewise linear approximations. Under this scenario, instead of considering continuous linear edges to determine the projection, instantaneous slopes of the curved edges are taken into account. Under such nonlinear projections, the the projection of $A_0'A_n'$ is assumed to be a spatial function, modeled as,
\begin{equation}
	\widetilde{A_0A_n} (x, y) = \sum_{i, j = 0, 0}^{i, j = 3, 3}a_{i,j} x^i y^j.
\end{equation}
This curve can be linearly approximated at a point $ (x_0, y_0) $ as 
\begin{equation}
	\label{eq:line_approx}
	\begin{split}
	\overbar{A_0A_n} (x, y)_0 \approx \widetilde{A_0A_n} (x_0, y_0)  + \dfrac{\partial \widetilde{A_0A_n}}{\partial x}  |_{(x_0, y_0)} (x - x_0) \\+ \dfrac{\partial \widetilde{A_0A_n}}{\partial y}  |_{(x_0, y_0)} (y - y_0).
	\end{split}
\end{equation}
Similar approximations can then be made on the projection of the opposite edge of the paper given by $ \widetilde{B_0B_n} (x, y) $. 
For brevity, $ \overbar{A_0A_n}(x,y)_i $ and $ \widetilde{A_0A_n}(x_i,y_i) $ are written as $ \overbar{A}_i $ and $ \widetilde{A}_i $ respectively. As such, the approximate version of the edge $ A_0A_n $ at a point $ (x_i,y_i) $ can be expressed as
\begin{equation}
\label{eq:edge-a}
\overbar{A}_i = \widetilde{A}_i + \dfrac{\partial \widetilde{A}_i}{\partial x}(x - x_0) + \dfrac{\partial \widetilde{A}_i}{\partial y}(y - y_0).
\end{equation}
Simplifying \eqref{eq:edge-a} yields
\begin{equation}
\label{eq:linear-edge-a}
y = -\dfrac{\partial \widetilde{A}_i}{\partial x} \dfrac{\partial y}{\partial \widetilde{A}_i}x + \left( \dfrac{\partial \widetilde{A}_i}{\partial x} \dfrac{\partial y}{\partial \widetilde{A}_i} x_0 + \dfrac{\partial y}{\partial \widetilde{A}_i}(\overbar{A}_i - \widetilde{A}_i) + y_0  \right)
\end{equation}
Just like \eqref{eq:edge-a}, the approximation of the opposite edge of the projected image of the paper can be written as 
\begin{equation}
\label{eq:edge-b}
\overbar{B}_i = \widetilde{B}_i + \dfrac{\partial \widetilde{B}_i}{\partial x}(x - x_0) + \dfrac{\partial \widetilde{B}_i}{\partial y}(y - y_0),
\end{equation}
and it can be simplified to
\begin{equation}
\label{eq:linear-edge-b}
y = -\dfrac{\partial \widetilde{B}_i}{\partial x} \dfrac{\partial y}{\partial \widetilde{B}_i}x + \left( \dfrac{\partial \widetilde{B}_i}{\partial x} \dfrac{\partial y}{\partial \widetilde{B}_i} x_0 + \dfrac{\partial y}{\partial \widetilde{B}_i}(\overbar{B}_i - \widetilde{B}_i) + y_0  \right).
\end{equation}
The instantaneous vanishing point ($ V_i $) for the $i$-th division is defined as the point of intersection of \eqref{eq:linear-edge-a} and \eqref{eq:linear-edge-b}. If $ \dfrac{\partial \widetilde{A}_i}{\partial x} \dfrac{\partial y}{\partial \widetilde{A}_i} \approxeq \dfrac{\partial \widetilde{B}_i}{\partial x} \dfrac{\partial y}{\partial \widetilde{B}_i} $, the corresponding segments are parallel (or almost parallel) and thus, the corresponding vanishing point is assumed to be at infinity. Otherwise, the coordinates for $ V_i $  on the projected plane can be calculated as,
\begin{equation}
V_i = \left(\dfrac{\mathcal{D}_i - \mathcal{C}_i}{\mathcal{A}_i - \mathcal{B}_i}, \dfrac{\mathcal{A}_i \mathcal{D}_i - \mathcal{B}_i\mathcal{C}_i}{\mathcal{A}_i - \mathcal{B}_i}\right),
\end{equation}
where, 

$ \mathcal{A}_i = -\dfrac{\partial \widetilde{A}_i}{\partial x} \dfrac{\partial y}{\partial \widetilde{A}_i} $, 

$ \mathcal{B}_i = -\dfrac{\partial \widetilde{B}_i}{\partial x} \dfrac{\partial y}{\partial \widetilde{B}_i} $, 

$ \mathcal{C}_i = \dfrac{\partial \widetilde{A}_i}{\partial x} \dfrac{\partial y}{\partial \widetilde{A}_i} x_0 + \dfrac{\partial y}{\partial \widetilde{A}_i}(\overbar{A}_i - \widetilde{A}_i) + y_0 $, and 

$ \mathcal{D}_i = \dfrac{\partial \widetilde{B}_i}{\partial x} \dfrac{\partial y}{\partial \widetilde{B}_i} x_0 + \dfrac{\partial y}{\partial \widetilde{B}_i}(\overbar{B}_i - \widetilde{B}_i) + y_0 $.\\

\begin{figure}
	\centering
	\subfloat[original]{\label{fig:ori_1}\includegraphics[width=0.45\linewidth]{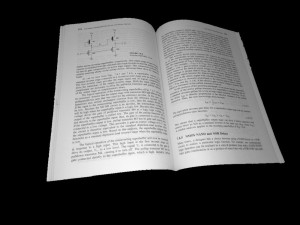}}\,
	\subfloat[curvilinear homography]{\label{fig:grid_1}\includegraphics[width=0.45\linewidth]{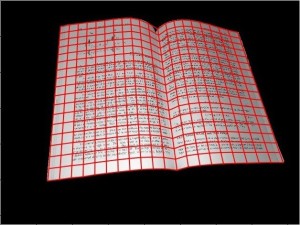}}\\
	\centering
	\subfloat[original]{\label{fig:ori_2}\includegraphics[width=0.45\linewidth]{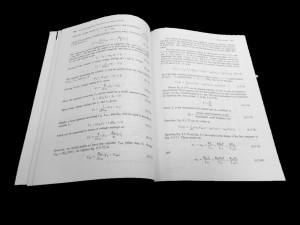}}\,
	\subfloat[curvilinear homography]{\label{fig:grid_2}\includegraphics[width=0.45\linewidth]{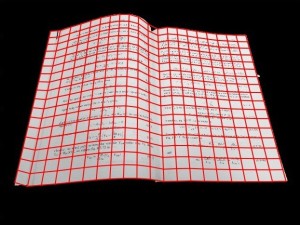}}\\
	\caption{\label{fig:grid-exaple-1} Depiction of grid generation on a document image under curvilinear homography. Notice that, the further the blocks are in the projected surface, the smaller they become.}
\end{figure}

Thus, if two opposite edges in an image of a paper under curvilinear projection are visible, a grid can be formed on the image that corresponds to equal areas on the unprojected plane. Fig. \ref{fig:grid-exaple-1} shows the formation of gridlines on one such image. The grid is designed in a manner such that, under an inverse perspective transformation, the blocks become squares of approximately the equal size. 
However, there can often be situations where the page boundaries are either unavailable or un-discernable. These might result from unintentional crops or premature preprocessing / binarization. Under these circumstances, the perspective transformation grid can be formed by recognizing two or more parallel printed lines on the warped document image and extrapolating from there. 
\subsection{Calculation of optimum inverse projection}
After the generation of the grid on the image based on curvilinear homography, the points of intersection of the gridlines are calculated. The objective is then to create  a normalization operator $ \bm{\tau}:\bm{v}\to \bm{w} $ for vectors $ \bm{v} $ and $ \bm{w} $ in $ \mathbb{R}^3 $. Here,  $ \bm{v} $ and $ \bm{w} $ are members of the vector space representing  the points in homogenous coordinates on the projected grid and the expected grid after inverse perspective transformation over the field $ \mathbb{R} $. Let the points and the centroid of the $ i $-th block on the projected grid be denoted by $ \bm{p}_{i, j} $ and $ C_i $ respectively. Let the normals on the gridlines in the $ i $-th block be denoted by $ \bm{N}_i $. Also, assume that $ \bm{D}_{i,j} $ represent the set of vectors directed from $ C_i $ to $ \bm{p}_{i, j} $. Let $ \bm{g}_{i,j} $ and $ \bm{h}_{i,j} $ represent the projected vectors in the direction of the major and minor directional flows (of the foreground objects, such as, the texts, images), respectively. Then, the average projection ($ \bm{P}_i $), major flow vector ($ \bm{G}_i $) and the minor flow vector ($ \bm{H}_i $) for the $ i $-th block can be calculated as
\begin{eqnarray}
\bm{P}_{i} &=& \sum_j \bm{\tau}\left\{ (\bm{p}_{i,j} \times \bm{D}_{i,j}) \times \bm{N}_i \right\},\nonumber\\
\bm{G}_{i} &=& \sum_j \bm{\tau}\left\{ (\bm{g}_{i,j} \times \bm{D}_{i,j}) \times \bm{N}_i \right\}, \textrm{ and},\\
\bm{H}_{i} &=& \sum_j \bm{\tau}\left\{ (\bm{h}_{i,j} \times \bm{D}_{i,j}) \times \bm{N}_i \right\}.\nonumber
\label{eq:projection-vec-calculation}
\end{eqnarray}
Here, the normalization operator $ \bm{\tau} $ can simply be assumed to be $ \bm{\tau}\{\bm{v}\}  = \dfrac{\bm{v}}{||\bm{v}||}$. 

Under the present scenario, $ \bm{P}_i $ can determine the average direction and magnitude of the projection vector of the $ i $-th block within the gridlines formed on the projected image.

The optimum inverse perspective projection for the $ i $-th block is then estimated by  performing a constrained optimization for $ \bm{N}_i $. It should also be noted that the optimizations performed on adjacent blocks must satisfy a smoothness criteria. This ensures that there are no abrupt changes in the inverse perspective transformations performed on adjacent blocks.

Let us assume that the average value of the major directional flow in the $ i $-th block is $ \bm{\bar{G}}_i $. Also, let $ \Delta $ denote the forward finite differences operator, for vector-valued functions. Now consider the following five constraints for the $ i $-th block:
\begin{enumerate}
	\item orthogonality between $ \bm{N}_i $ and $ \bm{P}_i $: $ \mu_{1,i} = ||\bm{N}_i \times \bm{P}_i||$;
	\item parallelism of the text-lines: $ \mu_{2,i} = ||\bm{G}_i - \bm{\bar{G}}_i|| $;
	\item geodesic property of the lines: $ \mu_{3, i} = ||\Delta \bm{\bar{G}}_i \cdot \bm{P}_i||^2 $;
	\item orthogonality of the text strokes and line directions: $ \mu_{4,i} = ||\bm{G}_i^T \cdot \bm{H}_i|| $;
	\item ensuring that the characters in the text-lines have the same height: $ \mu_{5,i} = \left|1-||\bm{N}_i||\, \right|^2 $.
\end{enumerate}
To calculate the optimum value of $ \bm{N}_i $ denoted by $ \bm{N}^*_i $, the weighted sum ($ \mathcal{Q}_i $) of the constraint functions is minimized. Thus, the objective is:
\begin{equation}
\min\limits_{\bm{N}_i} \mathcal{Q}_i(\bm{N}_i) = \min\limits_{\bm{N}_i} \sum_{k=1}^5 w_{k,i}\mu_{k,i} .
\label{eq:optimization}
\end{equation}
Through many trials on a standard dataset (CBDAR 2007 / IUPR 2011), it was found that the \textit{Truncated Generalized Lanczos algorithm (Krylov)} with a predefined trust region yielded the best compromise between the fastest and most accurate optimization for the objective \eqref{eq:optimization}. 

The process is started by initializing the vector $ \bm{\mu} = [\begin{array}{ccccc}
w_{1,i}\mu_{1,i} & w_{2,i}\mu_{2,i} &w_{3,i}\mu_{3,i} &w_{4,i}\mu_{4,i} &w_{5,i}\mu_{5,i} 
\end{array}]^T $ with fixed pre-defined values for $ w_{k,i} $ and random values for $ \mu_{k, i} $ within a certain bound. The optimization algorithm first approximates $ \mathcal{Q}(\bm{\mu}) $ as a quadratic function at the initial point $ \bm{\mu}_0 $ as
\begin{equation}
\begin{split}
\mathcal{Q}(\bm{\mu}) \approx  \mathcal{Q}(\bm{\mu}_0) + \bm{\nabla} \mathcal{Q}(\bm{\mu}_0) (\bm{\mu} - \bm{\mu}_0)\qquad \\\hfill+ \dfrac{1}{2}(\bm{\mu} - \bm{\mu}_0)^T \bm{\mathcal{H}}(\bm{\mu}_0) (\bm{\mu} - \bm{\mu}_0),
\end{split}
\end{equation}
where, $ \bm{\mathcal{H}}(\bm{\mu}_0) $ is the Hessian. If it is positive definite at $ \bm{\mu}_0 $, then, the local minimum of $ \mathcal{Q}(\bm{\mu}) $ can be calculated by setting $ \bm{\nabla} \mathcal{Q} = \bm{0} $. This yields,
\begin{equation}
\bm{\mu}_0^* = \bm{\mu}_0 - \bm{\mathcal{H}}\bm{\nabla} \mathcal{Q}.
\label{eq:hessian-inverse}
\end{equation}
The optimum $ \bm{\mu}_0 $ is calculated iteratively by using \eqref{eq:hessian-inverse}. First a maximum step size $ \delta_{max} $ is set and then the optimal step $ \bm{s} $ is calculated inside a given trust-region. For the $ k $-th step, the problem boils down to the following sub-problem:
\begin{equation}
\min\limits_{\bm{s}} \left\{ \mathcal{Q}(\bm{\mu}_k) + \bm{\nabla} \mathcal{Q}(\bm{\mu}_k) \cdot \bm{s} + \dfrac{1}{2}\bm{s}^T \bm{\mathcal{H}}(\bm{\mu}_k) \bm{s}  \right\},
\end{equation}
subject to: $ ||\bm{s}|| \leq \delta_{max} $.

The solution is updated in the next step as $ \bm{\mu}_{k+1} = \bm{\mu}_k + \bm{s} $. The trust region is then updated as per the desired accuracy of the solution. 

Once the optimum normals ($ \bm{N}_i $) for all the blocks are calculated in this manner, the mapping function for the inverse perspective transformation can be generated as a linear transformation that maps $ \bm{N}_i $ to $ \bm{N}_i^* $ for the $ i $-th block. 
\begin{figure}
	\centering
	\includegraphics[width=0.95\linewidth]{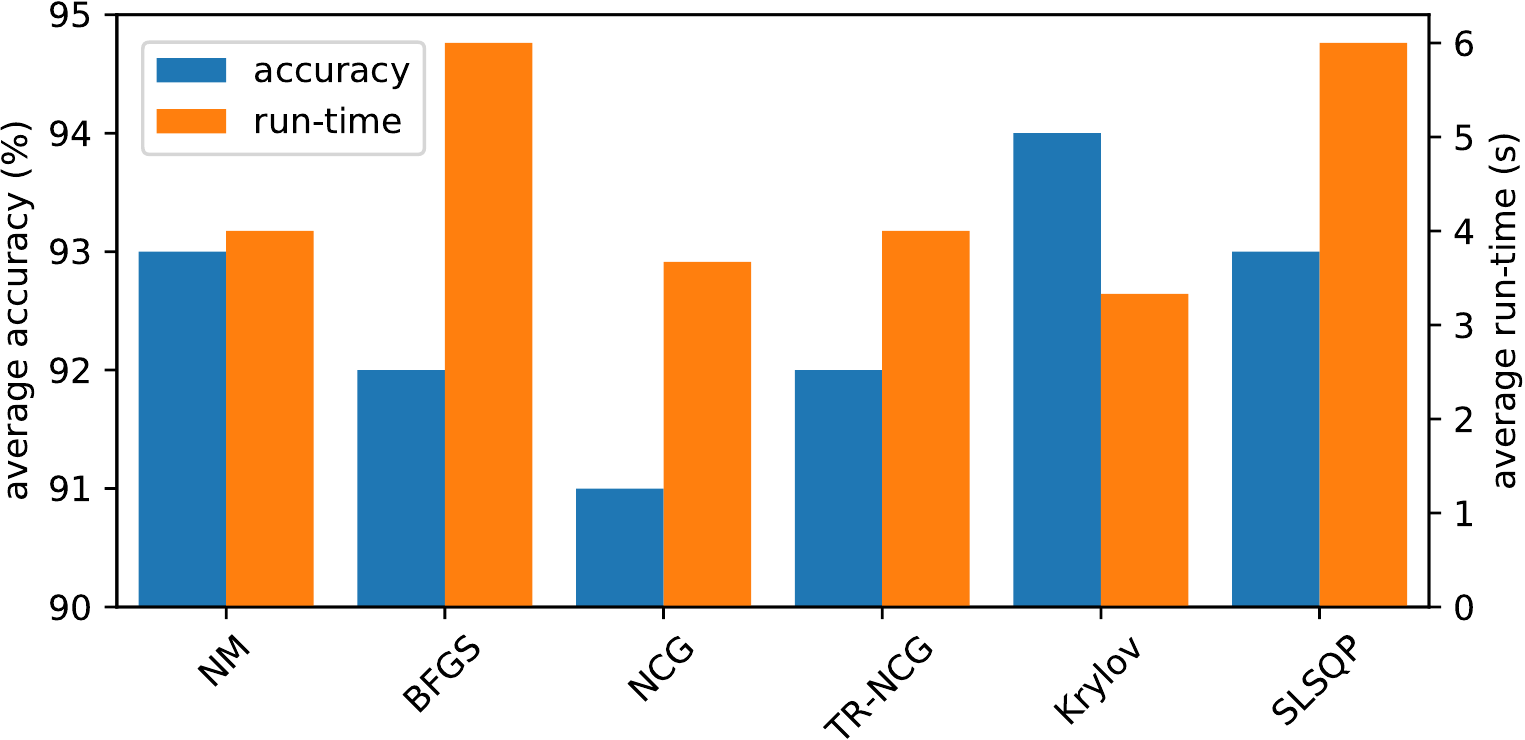}
	\caption{Comparison of different optimization techniques in the context of the present dewarping application. It was seen that the \textit{truncated generalized Lanczos algorithm (krylov)} performs the best in the present scenario. On an average, it provided the best optimization in the smallest run-time.}
	\label{fig:optimcomp}
\end{figure}
A comparison of the performance of few other optimization techniques which could have been used in this scenario is depicted in fig. \ref{fig:optimcomp}. Relevant optimization algorithms that are compared with the present one are: \textit{Nelder-Mead} (NM), \textit{Broyden-Fletcher-Goldfarb-Shanno} (BFGS), \textit{Newton-Conjugate-Gradient} (NCG), \textit{Trust-Region Newton-Conjugate-Gradient} (TR-NGC), and \textit{Sequential Least Squares Programming} (SLSQP). The accuracy is measured in terms of the OCR accuracy after one run of the page-level dewarping technique with the respective optimization algorithm.

\newcommand{\fcbdar}{0.16\linewidth}
\begin{figure*}
	\centering
	\subfloat[original]{\label{}\includegraphics[frame, height=\fcbdar]{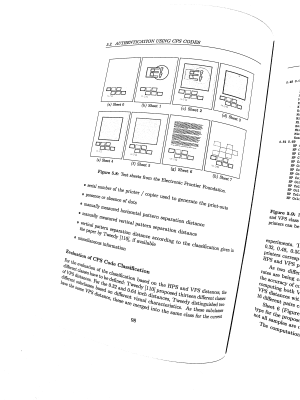}}
	\subfloat[dewarped]{\label{}\includegraphics[frame, height=\fcbdar]{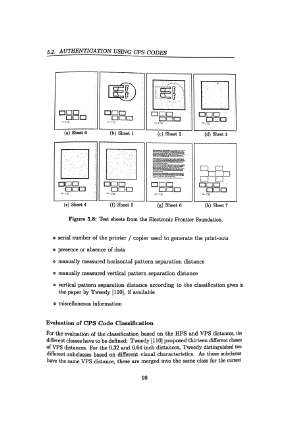}}\,
	\subfloat[original]{\label{}\includegraphics[frame, height=\fcbdar]{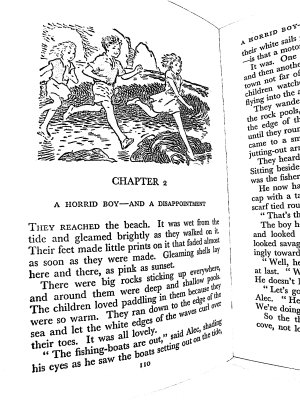}}
	\subfloat[dewarped]{\label{}\includegraphics[frame, height=\fcbdar]{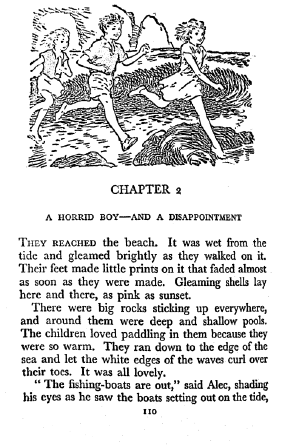}}\,
	\subfloat[original]{\label{}\includegraphics[frame, height=\fcbdar]{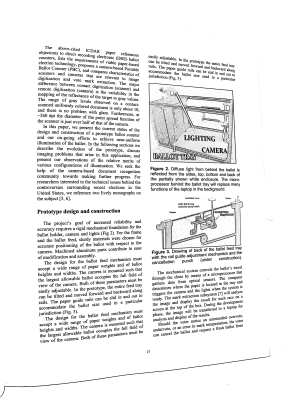}}
	\subfloat[dewarped]{\label{}\includegraphics[frame, height=\fcbdar]{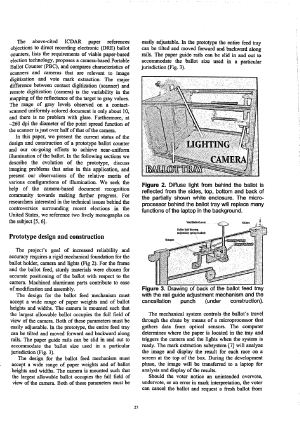}}\,
	\subfloat[original]{\label{}\includegraphics[frame, height=\fcbdar]{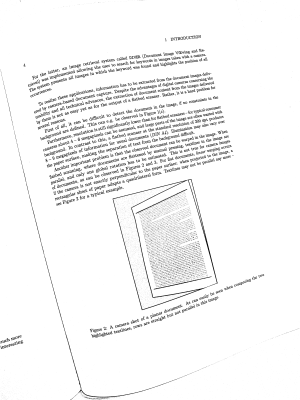}}
	\subfloat[dewarped]{\label{}\includegraphics[frame, height=\fcbdar]{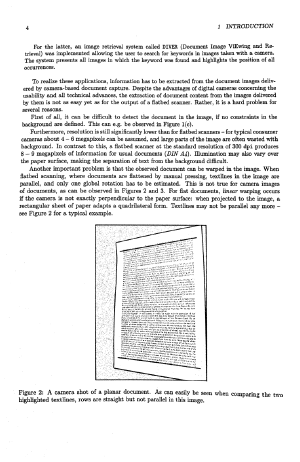}}\\
	\subfloat[original]{\label{}\includegraphics[frame, height=\fcbdar]{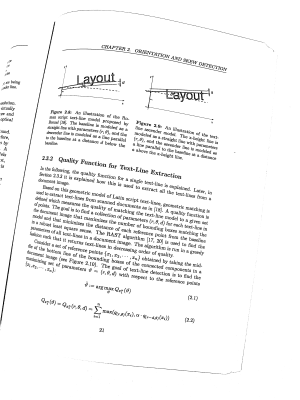}}
	\subfloat[dewarped]{\label{}\includegraphics[frame, height=\fcbdar]{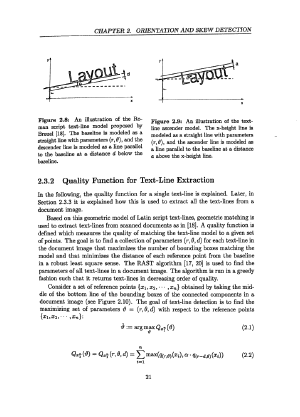}}\,
	\subfloat[original]{\label{}\includegraphics[frame, height=\fcbdar]{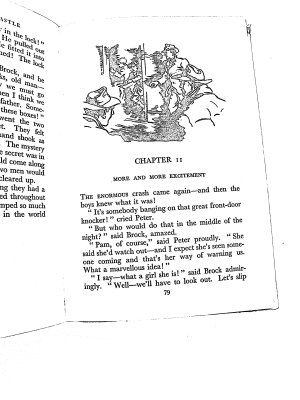}}
	\subfloat[dewarped]{\label{}\includegraphics[frame, height=\fcbdar]{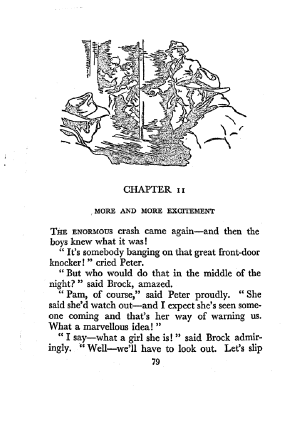}}\,
	\subfloat[original]{\label{}\includegraphics[frame, height=\fcbdar]{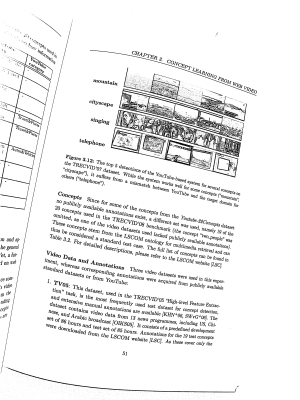}}
	\subfloat[dewarped]{\label{}\includegraphics[frame, height=\fcbdar]{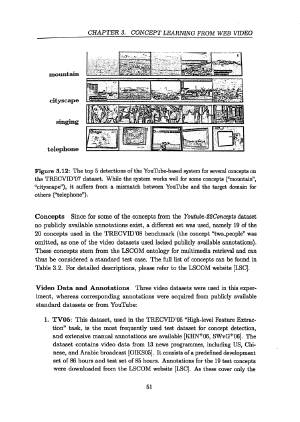}}\,
	\subfloat[original]{\label{}\includegraphics[frame, height=\fcbdar]{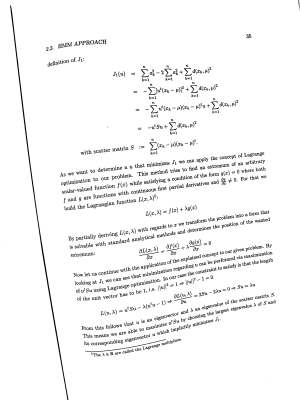}}
	\subfloat[dewarped]{\label{}\includegraphics[frame, height=\fcbdar]{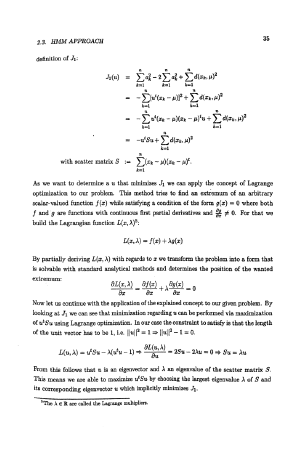}}
	\caption{\label{fig:first_pass} Sample results of  coordinate transform based dewarping of warped document images with optimum inverse projection. The tests are performed on CBDAR 2007 / IUPR 2011 document image dewarping  dataset. The samples shown here are chosen specifically to demonstrate the effectiveness of the procedure for various document layouts. These include: single column, double column, single column with images, double column with images and mathematical formulas.}
\end{figure*} 
The effectiveness of the described methodology is first tested on the 100 samples of the CBDAR 2007 / IUPR 2011 document image dewarping  dataset. A sample set of results are depicted in fig. \ref{fig:first_pass}. Notice that the page boundaries for these test images are not clearly visible. Thus, the homographic grid is constructed by correlating text-lines that are supposed to be parallel straight lines after the inverse perspective transformation. Embedded images, however, are a different challenge. In many documents, the embedded images are often ``framed". Such rectangular boundaries around images provide clear correlatable baselines for determining the homography. In other cases where such image boundaries are not available, the images are disregarded while estimating the homography. Once the homography for the text-lines are calculated, the grid is then extrapolated to the portions covered by the embedded images based on a presumed smoothness criteria. 

While forming perspective transformation grid on a warped document image, the average height of the text elements are estimated first. This is used as an initializer to determine a suitable step-size in the piecewise linear approximation of the curvilinear grid. It can be proved that the time complexity of the presented methodology is $ \mathcal{O}(n^3) $, where $ n $ is the number of sub-divisions, and as such, creating an unnecessarily dense grid would quickly result into a very large run-time. This approximation, however, sometimes leads to a dewarped image where the text-lines are not exactly parallel. This usually happens when the document image contains huge amount of unframed pictures and other unusual layouts. In any case, after dewarping, an  estimate is made for the loss of \textit{parallelism} and \textit{orthogonality} based on the methodology described earlier. If the loss is too high, the whole dewarping process is repeated with a denser grid. If the loss falls within an acceptable margin, individual text-lines are corrected with combined baseline / skew correction procedure. 
\subsection{Correction of nonlinear baselines and skews}
If a baseline / skew correction is deemed necessary, the image is first converted to a binary image where the foreground objects (text, images) are presented with white pixels and the background is presented in black. Subsequently, the image is denoised followed by \textit{morphological opening} with a small $ 3\times 3 $ or a $ 5 \times 5 $ square structuring element. Then, the resulting images are morphologically \textit{thinned} and \textit{pruned}. 

\begin{figure}
	\centering
	\subfloat[a sample text-line]{\label{}\includegraphics[frame, width=0.82\linewidth]{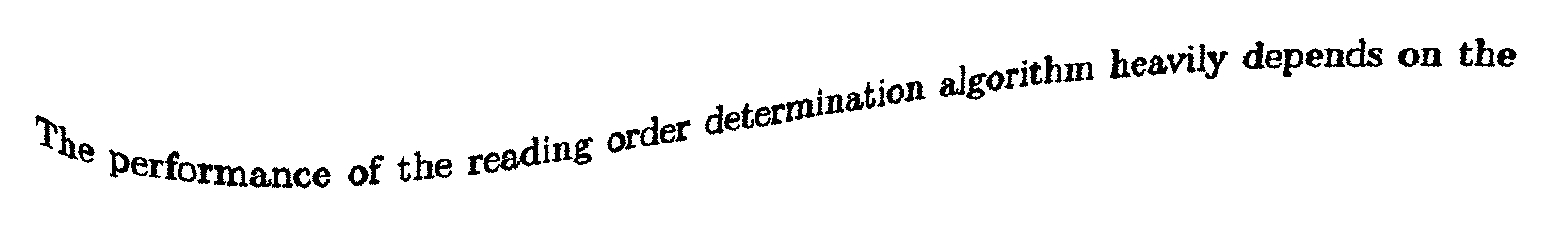}}\\
	\subfloat[after line level dewarping]{\label{}\includegraphics[width=0.86\linewidth]{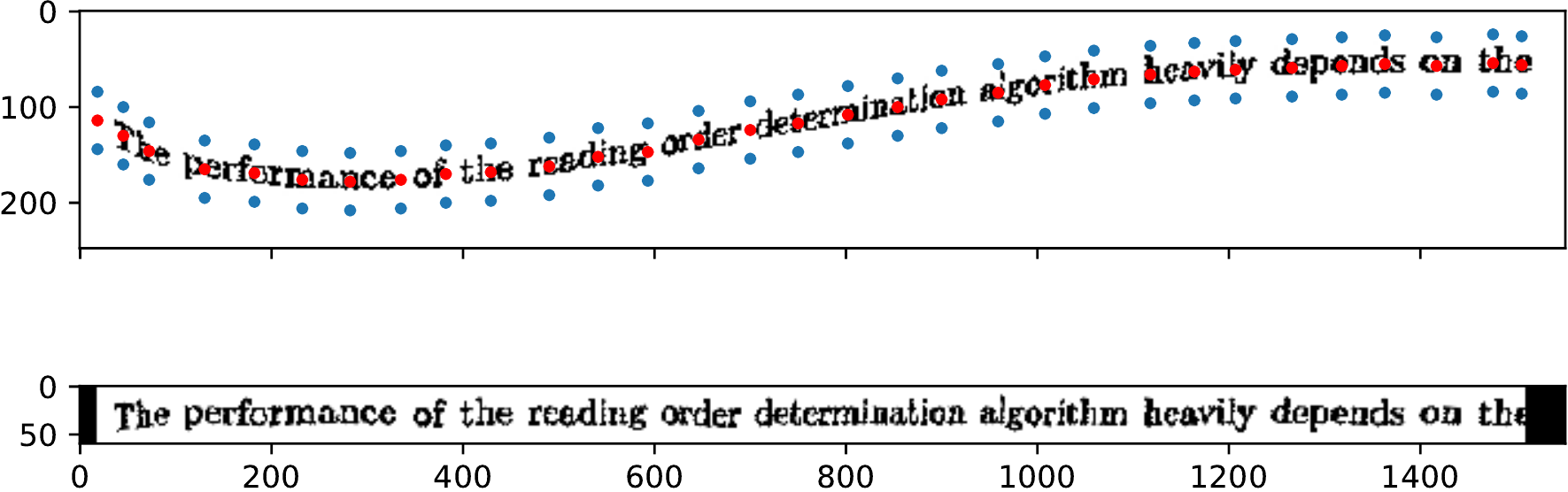}}\\
	\subfloat[a sample text-line]{\label{}\includegraphics[frame, width=0.82\linewidth]{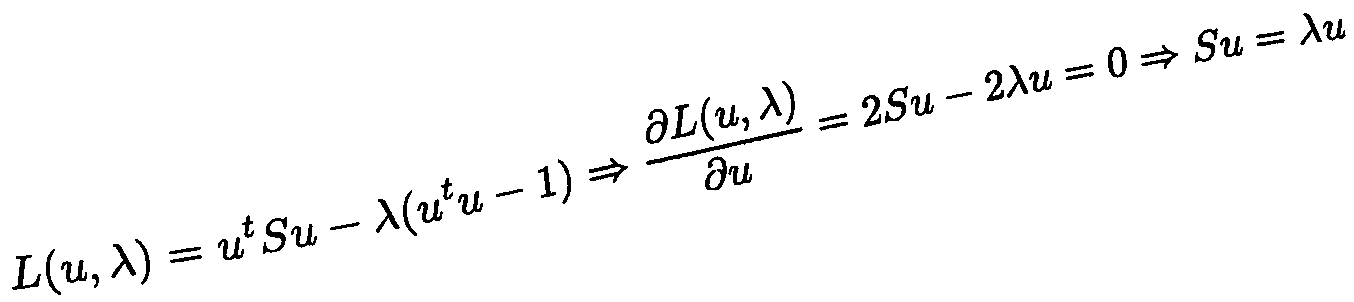}}\\
	\subfloat[after line level dewarping]{\label{}\includegraphics[width=0.9\linewidth]{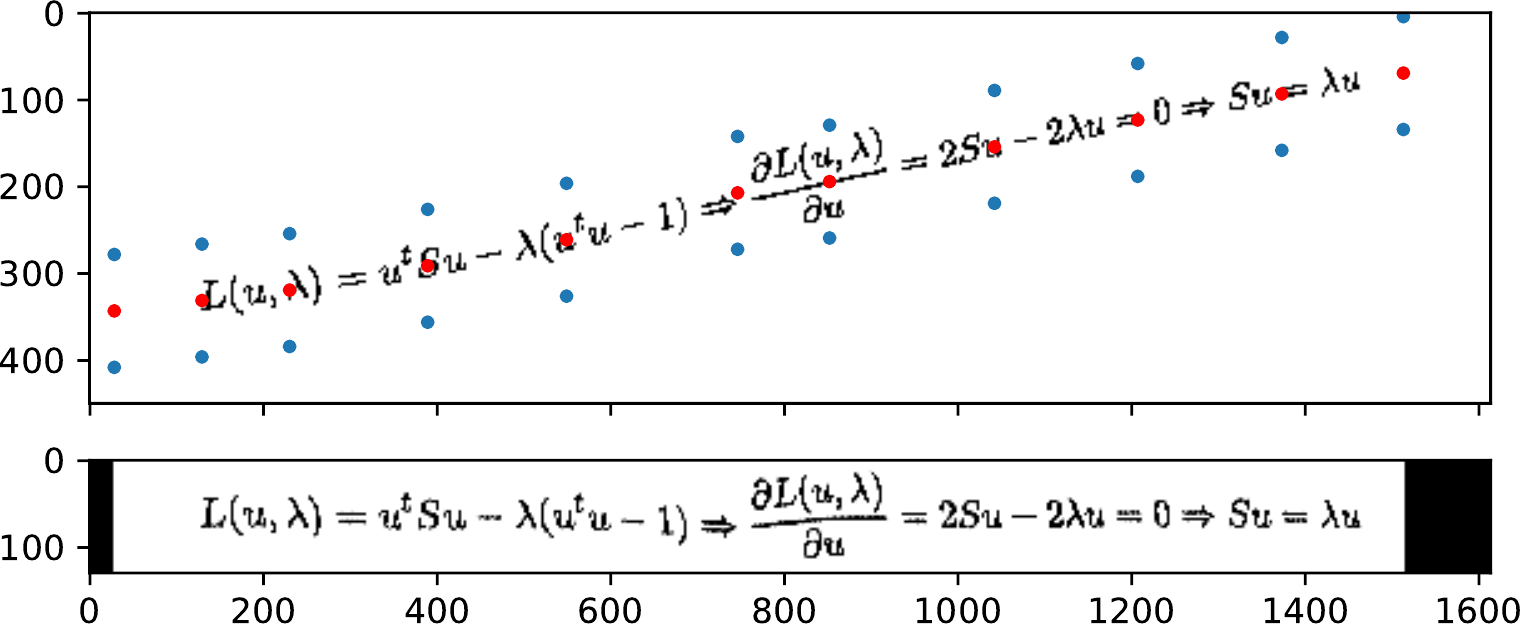}}\,
	\caption{\label{fig:line_level_correction} Depiction of line level dewarping through correction of nonlinear baselines and skews. (a) A sample text-line after suboptimal document level dewarping, and (b) dewarped through nonlinear baseline and skew correction. The red dots indicate the equally spaced points on the estimated baseline. The blue dots are the calculated points for generating the inverse affine transformation map. (c)--(d) Another example where the considered line is a mathematical equation.}
\end{figure}

Let $ \mathcal{A} $ be the pruned binary image. The skeletons produced through pruning would roughly represent the locations of the centroid of the text-lines. The objective is to generate smooth 8-connected paths representing the text-line without any intermediate breakage. For this, first a set of points $ \bm{x}_i, i=1, 2, \cdots, n $ on the pruned lines are selected at regular intervals satisfying a closeness criteria $ ||\bm{x}||< d_{nh} $. Then a cubic B-spline interpolation is generated as a curve $ \bm{x}(s) $, where the parameter $ s $ changes linearly within the points. The curves are then represented by $ \bm{x}(s) = \sum_{i=0}^{n+1}\bm{v}_i B_i(s) $, where $ B_i(s) $ is a cubic base function and $ \bm{v}_i $ are the spline coefficients. This is repeated for each of the detected text-lines on the image and splines $ \bm{\bar{x}}_k(s) $ representing each of the lines are generated. Regions of a small width $ d_{lw}  $ orthogonal to each spline is created iteratively to make an estimate for the width of each test line. The selected width is gradually increased until the selected regions completely engulf the text-lines under consideration. Each of the splines are then re-sampled at regular intervals to generate baselines for each individual text-line. If $ (x_i,y_i) $ and $ (x_{i+1}, y_{i+1}) $ are two such consecutive points on a text-line, the instantaneous principle slope angle of the text-line is determined as
$$ \theta_i = \tan^{-1}\left(\left| \frac{y_{i+1}-y_i}{x_{i+1} - x_i} \right|\right).$$ The midpoint is represented by $ (\bar{x}_i,\bar{y}_i) = ((x_{i+1} + x_i)/2, \, (y_{i+1} + y_i)/2) $. The points $ (x_{ui}, y_{ui}) $ and $ (x_{li},y_{li}) $ on the upper and lower boundaries, respectively, of the orthogonal neighborhood of the part of the text-line between $ (x_i,y_i) $ and $ (x_{i+1}, y_{i+1}) $ are then represented by 
\begin{equation}
\begin{split}
	x_{ui} = \lfloor \bar{x}_i - d_{lw} \sin(\theta_i) \rfloor,\quad
	y_{ui} = \lfloor \bar{y}_i + d_{lw} \cos(\theta_i) \rfloor;\\
	x_{li} = \lfloor \bar{x}_i + d_{lw} \sin(\theta_i) \rfloor,\quad
	y_{li} = \lfloor \bar{y}_i - d_{lw} \cos(\theta_i) \rfloor.
\end{split}
\label{eq:boundary}
\end{equation}
The simplified version of the piecewise affine transform for dewarping the text-lines is given by $f:U\to V$ where $U$ and $V$ are affine spaces and $f$ is of the form $x \mapsto Gx + h$, where $G$ is a linear transformation over $U$ and $h \in V$.
The transformation can be rewritten as
\begin{equation}\label{eq:affine}
\begin{bmatrix} \bm{Y} \\ 1 \end{bmatrix} = \left[ \begin{array}{ccc|c} \, & \bm{G} & & \ \bm{H} \ \\ 0 & \ldots & 0 & 1 \end{array} \right] \begin{bmatrix} \bm{X} \\ 1 \end{bmatrix}.
\end{equation}
This can be further simplified to
\begin{equation}
\begin{split}
x_i'=x_i\cos\phi-y_i\sin\phi+c, \\ y_i'=y_i\cos\phi+y_i\sin\phi +f.
\end{split}
\label{eq:affine2}
\end{equation}
The transformation from a set of points $(x_i,y_i)$ to a projected set of points $(x_i', y_i')$ is calculated as
\begin{equation}
(x_i',y_i') =\left(\frac{a_{0,i} x_i + a_{1,i} y_i + a_{2,i}}{c_{0,i} x_i + c_{1,i} y_i + 1}, \frac{b_{0,i} x_i + b_{1,i} y_i + b_{2,i}}{c_{0,i} x_i + c_{1,i} y_i + 1}\right).
\label{eq:affine3}
\end{equation}
Putting \eqref{eq:affine3} in \eqref{eq:affine} and setting coefficients $ c_{0,i} = 0 $ and $ c_{1,i} = 0 $, the affine transformation matrix $\bm{G}$ becomes, 
$$ \bm{G} =
\left[\begin{array}{ccc} \, a_{0,i} & a_{1,i} & a_{2,i} \\ b_{0,i} & b_{1,i} & b_{2,i} \\ 0 & 0 & 1 \end{array}\right].
$$
In general, the affine transformation from $(x,y)$ to $(x', y')$ is given by
\begin{equation} \label{eq:affine5}
\begin{aligned}
&\bm{x}' = \bm{Sc}_x \bm{x}\cos(\bm{Ro}) - \bm{Sc}_y \bm{y}\sin(\bm{Ro} + \bm{Sh}) + \bm{Tr}_x \\
&\bm{y}' = \bm{Sc}_x \bm{x}\sin(\bm{Ro}) - \bm{Sc}_y \bm{y}\cos(\bm{Ro} + \bm{Sh}) + \bm{Tr}_y
\end{aligned}.
\end{equation}
where, $ \bm{Sc} $, $ \bm{Ro} $, $ \bm{Sh} $ and $ \bm{Tr} $ denote scale, rotation, shear and translation, respectively.
\begin{figure}
	\centering
	\subfloat[original]{\label{}\includegraphics[frame, height=0.4\linewidth]{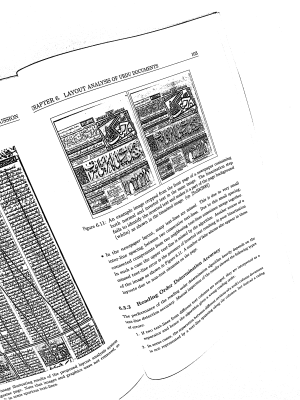}}\,
	\subfloat[suboptimally dewarped]{\label{}\includegraphics[frame, height=0.4\linewidth]{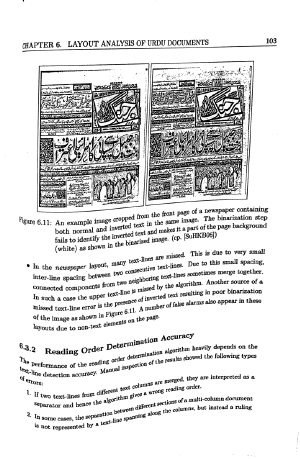}}\,
	\subfloat[optimally dewarped]{\label{}\includegraphics[frame, height=0.4\linewidth]{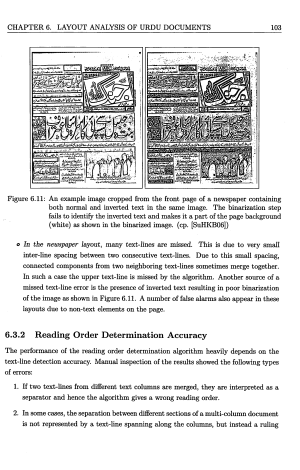}}\,
	\caption{\label{fig:second_pass} An example of suboptimal result obtained when the document layout is too complicated}
\end{figure} 
Fig. \ref{fig:second_pass} shows an example where the first pass of the page level dewarping yields unsatisfactory result. This is mainly due to the fact that the layout of the document is just too complicated for the presented technique. To combat that, the suboptimally dewarped image is then passed through the line level dewarping process that corrects the nonlinear baselines and skews in individual text-lines. The process of line-level dewarping on one such line is depicted in fig. \ref{fig:line_level_correction}.
\section{Experimental Results}
The effectiveness of the present methodology is first assessed by performing tests on the CBDAR 2007 / IUPR 2011 document image dewarping  dataset. This dataset is considered to be the \textit{de-facto} standard for testing document image dewarping techniques, since, it was designed with a multitude of standard use-cases in mind. The results are compared with various other well-established and state-of-the-art techniques in the domain. Table \ref{tab:techniques} provides a brief overview of the selected techniques.
\begin{table}[h]
	\centering 
	\rowcolors{2}{white}{gray!25}
	\caption{List of techniques with which the present methodology is compared\label{tab:techniques}}
	\begin{tabular*}{\linewidth}{p{70pt}p{159pt}}    \toprule
		\textbf{algorithm} & \textbf{based on} \\\midrule
		Masalovitch \cite{masalovitch2007usage}  & continuous skeletal image representation \\
		Gatos \cite{Gatos}					& segmentation of text-lines and words \\
		Stamatopoulos \cite{5585760}		& goal-oriented coarse to fine corrections \\
		Wu \cite{Wu_Model}					& 2d coordinate transform model\\
		Bukhari \cite{Bukhari}				& coupled snakelet model for curled text-lines\\
		Meng \cite{Meng_Metric}				& metric rectification\\
		Kil \cite{Kil}						& recognition of text-lines and line segments\\
		Ma \cite{Ma_DocUNet} 				& deep convolutional stacked u-nets\\
		\hline
	\end{tabular*}
\end{table}
However, before dewarping the images, often a pre-processing step can be employed to discard regions outside the expected regions of interest (ROI). In the IUPR 2011 dataset, both greyscale and binarized images samples are available. In such situations, page boundaries and the corresponding ROIs can be calculated using a combination of watershed-based segmentation, connectivity-based component analysis and clustering. 
\subsection{Page segmentation and ROI extraction from grayscale images}
After performing a contrast enhancement \cite{Adhikari_histogram_correction} on the collected image, the image is re-scaled to a smaller size and converted to a grayscale one. The resulting image is then segmented using the classical watershed technique. This produces labels for different regions in the image, providing a correlation between different pixel intensity levels in the image. The set of connected components are then obtained and small clusters are discarded. Thus the ROI turns out to be in one of the larger clusters. A hull (which is not necessarily convex) is generated over this cluster through a morphological hole-filling algorithm. This hull is then used as a mask to extract the ROI from the image. The whole process is depicted in algorithm \ref{algo:ROI} and an example is shown in fig. \ref{fig:ROI}. 
\newcommand{\froi}{0.25\linewidth}
\begin{figure}
	\centering
	\subfloat[]{\label{}\includegraphics[frame, height=\froi]{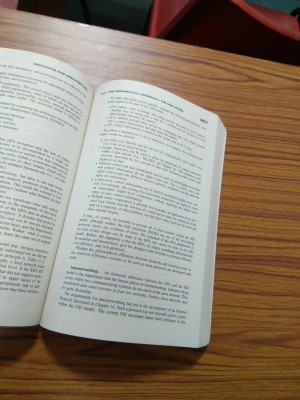}}
	\subfloat[]{\label{}\includegraphics[frame, height=\froi]{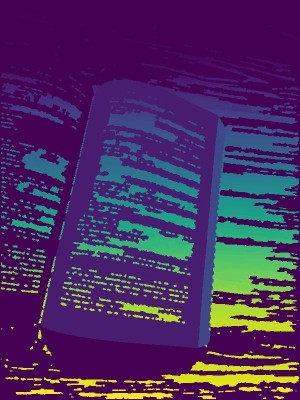}}
	\subfloat[]{\label{}\includegraphics[frame, height=\froi]{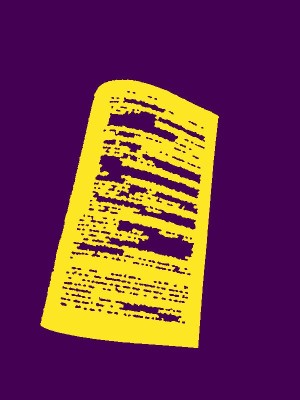}}
	\subfloat[]{\label{}\includegraphics[frame, height=\froi]{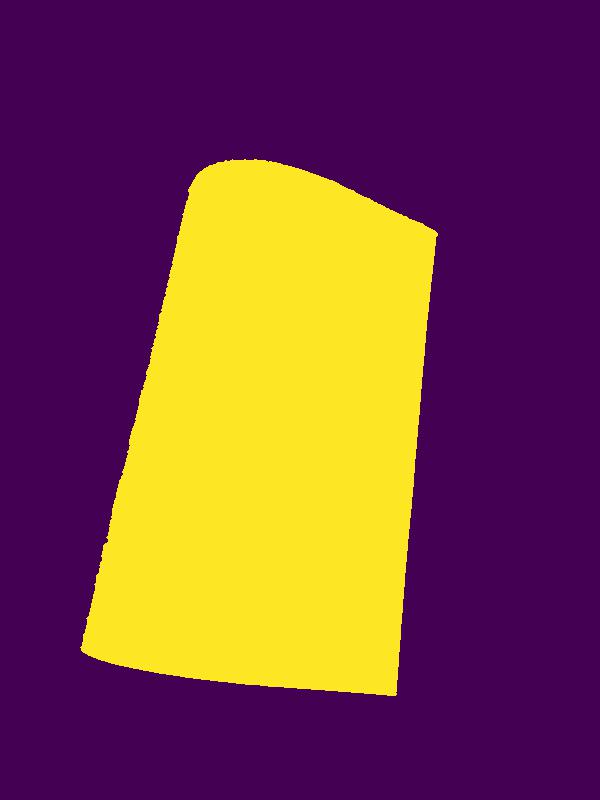}}
	\subfloat[]{\label{}\includegraphics[frame, height=\froi]{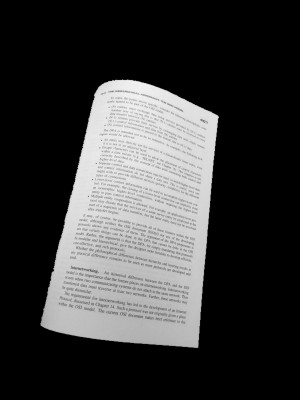}}
	\caption{\label{fig:ROI} An example of page boundary and ROI extraction. (a) A warped document image taken with a handheld camera; (b) after application of watershed segmentation; (c) extraction of connected components through k-means clustering; (d) hull generation and (e) page segmentation. }
\end{figure}

\begin{algorithm}
	\SetAlgoLined
	\SetKwInOut{Output}{Output}
	\KwIn{Warped document image: $ \bm{A} $\;
		Minimum expected foreground coverage as a rough estimate of the amount of image area covered by the document:  $ h_{min} $\;}
	\Output{Segmented grayscale image $ \hat{\bm{A}} $ containing only the ROI\;}
	\Begin
	{	
		Initialize a scaling factor $ s \leftarrow 1 $\;
		Initialize a hull-ratio $ h_r \leftarrow 0$\;
		Set a max target dimension $ d_{max} $\;
		Set $ r \leftarrow \mathrm{row\_dim}(\bm{A})$ and $ c \leftarrow \mathrm{col\_dim}(\bm{A})$\;
		\If{$ r >  d_{max}$ and $ c >  d_{max}$}
		{
			$ s \leftarrow d_{max} / \max(r, d) $\;
		}
		$ \bar{\bm{A}} \leftarrow \mathrm{image\_rescale}(\bm{A},\, \mathrm{scale}=s) $\;
		$ {\bm{A}_h} \leftarrow \mathrm{correct\_histogram}(\bar{\bm{A}})$\;
		$ {\bm{A}_g} \leftarrow \mathrm{convert\_to\_gray}(\bm{A}_h)$\;
		$ {\bm{A}_p} \leftarrow \mathrm{scharr}(\bm{A}_g)$ /* Scharr filter for extracting outlines */\;
		$ {\bm{T}} \leftarrow \mathrm{otsu\_threshold}(\bm{A}_p)$ /* Otsu threshold value for each pixel in $ \bm{A}_p $ */\; 
		Calculate the binary image outline: $ \bm{A}_{bo} \leftarrow \mathrm{bool}(\bm{A}_p \geq \bm{T}) $\;
		Set initial watershed compactness $ c \leftarrow 1 $
		\Repeat{$ h_r > h_{min} $}
		{
			Perform watershed segmentation: $ \bm{A}_{w} \leftarrow $ $\mathrm{watershed}(\bm{A}_{bo}, \, \mathrm{compactness}=c)$\;
			$ n_w \leftarrow \mathrm{bincount}(\bm{A}_{w}) $ /* count the number of distinct levels */\;
			Calculate the most common levels: $ \bm{L}_{C} \leftarrow \mathrm{argpartition}(-n_w)$\;
			\For{level $ L $ in $ \bm{L}_{C} $}
			{
				$ \bm{A}_{gw} \leftarrow \mathrm{bool}(\bm{A}_{w} == L) $ /* segment based on the most common level values */\;
				$ \bm{A}_{H} \leftarrow \mathrm{binary\_fill\_holes}(\bm{A}_{gw})$ /* generate hull image (the hull is not necessarily convex) */\;
				Find the number of objects in the hull: $ n_{obj} \leftarrow \mathrm{count\_unique\_elements}(\bm{A}_{H}) $ /* the intended one would have only two objects in it */\;
				\If{$ n_{obj} == 2$}{\textbf{\textit{break}}\;}
			}
			Recalculate hull-ratio: $ h_r \leftarrow \dfrac{\mathrm{sum}(\bm{A}_{H})}{s^2rc}$\;
			Set new compactness: $ c \leftarrow 10 c $\;
		}
		Resize the hull back to the original size: $ \bm{A}_{H} \leftarrow \mathrm{image\_rescale}(\bm{A}_H,\, \mathrm{scale}=1/s) $\;
		Generate the segmented page: $ \hat{\bm{A}} \leftarrow \mathrm{logical\_and}(\bm{A},\,\bm{A}_{H})$\;
		\Return $ \hat{\bm{A}} $\;
		\caption{Extraction of ROI from warped document images\label{algo:ROI}}
	}
\end{algorithm}

\subsection{Dewarping}
\begin{table*}
	\centering 
	\rowcolors{2}{white}{gray!25}
	\caption{Depiction of the quality assessment and decision making process after one application of page-level dewarping. After a single pass of page-level dewarping, further decisions are made based on the obtained value of the average value of the quality metric $ \tilde{\mu} $.  \label{tab:step-by-step}}
	\begin{tabular*}{\linewidth}{rcrccccccr|rcrccccccr}    \toprule
		\textbf{sl} & \textbf{acc} & \textbf{im} & $ \overbar{\mu}_1 $ &  $ \overbar{\mu}_2 $ &  $ \overbar{\mu}_3 $ &  $ \overbar{\mu}_4 $ &  $ \overbar{\mu}_5 $ & $ \tilde{\mu} $ & \textbf{dec} & \textbf{sl} & \textbf{acc} & \textbf{im} & $ \overbar{\mu}_1 $ &  $ \overbar{\mu}_2 $ &  $ \overbar{\mu}_3 $ &  $ \overbar{\mu}_4 $ &  $ \overbar{\mu}_5 $ & $ \tilde{\mu} $ & \textbf{dec}\\\midrule
		1 & 79.27 & 65 & 0.84 & 0.31 & 0.67 & 0.71 & 0.14 & 0.75 & R+L & 51 & 89.31 & 22 & 1.00 & 0.00 & 0.78 & 0.78 & 0.04 & 0.90 & L\\
		2 & 81.58 & 52 & 0.76 & 0.27 & 0.99 & 0.73 & 0.20 & 0.80 & R+L & 52 & 89.43 & 61 & 0.82 & 0.00 & 0.85 & 0.93 & 0.09 & 0.90 & L\\ 
		3 & 81.61 & 83 & 0.83 & 0.22 & 0.96 & 0.88 & 0.24 & 0.84 & R+L & 53 & 89.56 & 96 & 1.00 & 0.02 & 0.95 & 0.88 & 0.16 & 0.93 & L\\
		4 & 82.32 & 13 & 0.75 & 0.22 & 0.72 & 0.78 & 0.14 & 0.78 & R+L & 54 & 89.66 & 80 & 0.95 & 0.10 & 0.92 & 0.88 & 0.15 & 0.90 & L\\ 
		5 & 82.53 & 71 & 0.88 & 0.00 & 0.96 & 0.98 & 0.17 & 0.93 & L & 55 & 89.72 & 89 & 1.00 & 0.05 & 0.79 & 0.93 & 0.23 & 0.89 & R+L\\
		6 & 82.76 & 18 & 0.80 & 0.17 & 0.77 & 0.78 & 0.00 & 0.84 & R+L & 56 & 89.72 & 27 & 0.86 & 0.07 & 0.76 & 1.00 & 0.01 & 0.91 & L\\
		7 & 83.63 & 50 & 0.83 & 0.00 & 0.80 & 0.76 & 0.00 & 0.88 & R+L & 57 & 89.81 & 75 & 1.00 & 0.21 & 0.84 & 0.95 & 0.00 & 0.92 & L\\
		8 & 83.99 & 20 & 0.88 & 0.26 & 0.71 & 0.99 & 0.00 & 0.86 & R+L & 58 & 89.87 & 9 & 0.90 & 0.06 & 1.00 & 0.90 & 0.17 & 0.91 & L\\ 
		9 & 84.22 & 42 & 0.79 & 0.06 & 0.86 & 0.95 & 0.03 & 0.90 & L & 59 & 89.99 & 82 & 1.00 & 0.06 & 0.81 & 1.00 & 0.22 & 0.91 & L\\
		10 & 84.39 & 28 & 0.91 & 0.14 & 0.86 & 0.95 & 0.06 & 0.90 & L & 60 & 90.11 & 17 & 1.00 & 0.13 & 0.88 & 0.87 & 0.00 & 0.92 & L\\ 
		11 & 84.56 & 41 & 0.78 & 0.23 & 0.77 & 0.89 & 0.29 & 0.78 & R+L & 61 & 90.19 & 1 & 0.90 & 0.14 & 0.86 & 0.99 & 0.10 & 0.90 & L\\
		12 & 84.88 & 55 & 0.77 & 0.00 & 0.71 & 0.82 & 0.17 & 0.83 & R+L & 62 & 90.48 & 60 & 0.88 & 0.05 & 0.92 & 0.90 & 0.12 & 0.91 & L\\ 
		13 & 85.08 & 12 & 0.89 & 0.09 & 0.74 & 0.88 & 0.06 & 0.87 & R+L & 63 & 90.50 & 48 & 0.79 & 0.00 & 0.83 & 1.00 & 0.10 & 0.90 & L\\ 
		14 & 85.42 & 40 & 0.84 & 0.24 & 0.77 & 1.00 & 0.21 & 0.83 & R+L & 64 & 90.52 & 44 & 0.93 & 0.00 & 0.86 & 0.96 & 0.04 & 0.94 & L\\ 
		15 & 85.52 & 86 & 0.81 & 0.13 & 0.85 & 0.72 & 0.22 & 0.81 & R+L & 65 & 90.54 & 11 & 0.97 & 0.07 & 0.86 & 0.90 & 0.16 & 0.90 & L\\
		16 & 85.59 & 98 & 0.72 & 0.04 & 0.94 & 0.79 & 0.16 & 0.85 & R+L & 66 & 90.55 & 10 & 1.00 & 0.03 & 0.89 & 0.85 & 0.15 & 0.91 & L\\ 
		17 & 85.67 & 97 & 0.83 & 0.11 & 0.88 & 0.84 & 0.00 & 0.89 & R+L & 67 & 90.73 & 21 & 1.00 & 0.23 & 1.00 & 0.93 & 0.00 & 0.94 & L\\
		18 & 85.91 & 15 & 0.91 & 0.20 & 0.86 & 0.76 & 0.00 & 0.87 & R+L & 68 & 90.77 & 59 & 1.00 & 0.00 & 0.78 & 0.76 & 0.00 & 0.91 & L\\
		19 & 86.08 & 49 & 1.00 & 0.10 & 0.74 & 1.00 & 0.26 & 0.88 & R+L & 69 & 90.94 & 16 & 0.92 & 0.14 & 0.99 & 0.86 & 0.10 & 0.91 & L\\
		20 & 86.53 & 79 & 1.00 & 0.03 & 1.00 & 0.85 & 0.19 & 0.93 & L & 70 & 91.12 & 73 & 0.81 & 0.00 & 0.91 & 0.85 & 0.07 & 0.90 & L\\ 
		21 & 86.61 & 88 & 0.75 & 0.26 & 1.00 & 0.88 & 0.08 & 0.86 & R+L & 71 & 91.27 & 70 & 1.00 & 0.20 & 0.91 & 0.86 & 0.00 & 0.91 & L\\
		22 & 86.94 & 62 & 0.87 & 0.18 & 0.92 & 0.79 & 0.21 & 0.84 & R+L & 72 & 91.34 & 69 & 0.85 & 0.08 & 0.90 & 0.88 & 0.01 & 0.91 & L\\
		23 & 86.97 & 51 & 0.89 & 0.26 & 1.00 & 0.86 & 0.02 & 0.89 & R+L & 73 & 91.54 & 23 & 0.85 & 0.03 & 0.90 & 0.92 & 0.14 & 0.90 & L\\
		24 & 87.11 & 72 & 1.00 & 0.22 & 0.98 & 0.78 & 0.21 & 0.87 & R+L & 74 & 91.59 & 32 & 0.92 & 0.20 & 0.87 & 0.78 & 0.19 & 0.84 & R+L\\ 
		25 & 87.24 & 53 & 1.00 & 0.18 & 0.88 & 1.00 & 0.23 & 0.89 & R+L & 75 & 91.66 & 3 & 0.89 & 0.00 & 1.00 & 1.00 & 0.23 & 0.93 & L\\
		26 & 87.27 & 84 & 0.89 & 0.00 & 0.91 & 0.83 & 0.11 & 0.90 & L & 76 & 91.70 & 25 & 0.89 & 0.12 & 0.91 & 0.95 & 0.11 & 0.90 & L\\ 
		27 & 87.31 & 77 & 0.75 & 0.00 & 1.00 & 0.85 & 0.16 & 0.89 & R+L & 77 & 91.74 & 7 & 1.00 & 0.18 & 0.86 & 0.96 & 0.08 & 0.91 & L\\
		28 & 87.35 & 78 & 0.86 & 0.13 & 0.91 & 1.00 & 0.13 & 0.90 & L & 78 & 91.74 & 66 & 0.99 & 0.06 & 0.88 & 0.92 & 0.04 & 0.94 & L\\ 
		29 & 87.37 & 36 & 1.00 & 0.14 & 0.91 & 1.00 & 0.20 & 0.91 & L & 79 & 91.82 & 31 & 0.82 & 0.06 & 0.97 & 0.90 & 0.00 & 0.93 & L\\
		30 & 87.41 & 81 & 0.84 & 0.01 & 1.00 & 0.95 & 0.08 & 0.94 & L & 80 & 92.08 & 64 & 1.00 & 0.00 & 0.86 & 0.90 & 0.13 & 0.93 & L\\ 
		31 & 87.51 & 39 & 0.98 & 0.14 & 0.81 & 0.87 & 0.00 & 0.90 & L & 81 & 92.26 & 19 & 0.86 & 0.00 & 0.90 & 1.00 & 0.13 & 0.93 & L\\
		32 & 87.51 & 76 & 0.83 & 0.00 & 0.90 & 0.88 & 0.26 & 0.87 & R+L & 82 & 92.33 & 47 & 1.00 & 0.22 & 1.00 & 0.96 & 0.20 & 0.91 & L\\ 
		33 & 87.54 & 5 & 0.81 & 0.17 & 0.91 & 1.00 & 0.00 & 0.91 & L & 83 & 92.35 & 8 & 0.94 & 0.09 & 0.93 & 0.82 & 0.00 & 0.92 & L\\
		34 & 87.57 & 90 & 0.89 & 0.04 & 0.91 & 0.92 & 0.16 & 0.90 & L & 84 & 92.51 & 4 & 0.91 & 0.00 & 0.89 & 0.94 & 0.05 & 0.94 & L\\ 
		35 & 87.59 & 92 & 0.82 & 0.14 & 0.92 & 0.76 & 0.24 & 0.82 & R+L & 85 & 92.57 & 30 & 1.00 & 0.00 & 0.84 & 0.92 & 0.15 & 0.92 & L\\
		36 & 87.70 & 35 & 0.96 & 0.09 & 1.00 & 0.83 & 0.00 & 0.94 & L & 86 & 92.68 & 100 & 0.81 & 0.07 & 1.00 & 0.91 & 0.11 & 0.91 & L\\ 
		37 & 87.74 & 54 & 0.84 & 0.00 & 0.97 & 0.89 & 0.16 & 0.91 & L & 87 & 92.87 & 93 & 0.81 & 0.20 & 1.00 & 1.00 & 0.03 & 0.92 & L\\
		38 & 88.23 & 99 & 0.93 & 0.00 & 0.86 & 0.90 & 0.10 & 0.92 & L & 88 & 92.90 & 45 & 0.94 & 0.00 & 0.84 & 0.87 & 0.00 & 0.93 & L\\ 
		39 & 88.29 & 37 & 1.00 & 0.00 & 0.95 & 0.88 & 0.24 & 0.92 & L & 89 & 93.18 & 57 & 0.94 & 0.02 & 0.91 & 0.98 & 0.14 & 0.93 & L\\
		40 & 88.40 & 33 & 1.00 & 0.25 & 0.88 & 0.84 & 0.17 & 0.86 & R+L & 90 & 93.26 & 91 & 0.81 & 0.06 & 0.82 & 1.00 & 0.00 & 0.91 & L\\ 
		41 & 88.43 & 38 & 0.82 & 0.10 & 0.93 & 0.88 & 0.02 & 0.90 & L & 91 & 93.38 & 43 & 1.00 & 0.00 & 0.84 & 0.81 & 0.00 & 0.93 & L\\
		42 & 88.68 & 74 & 0.87 & 0.07 & 0.86 & 0.90 & 0.00 & 0.91 & L & 92 & 93.79 & 56 & 0.93 & 0.00 & 0.85 & 0.91 & 0.08 & 0.92 & L\\ 
		43 & 88.77 & 85 & 0.95 & 0.15 & 0.84 & 0.91 & 0.03 & 0.90 & L & 93 & 94.54 & 6 & 1.00 & 0.14 & 0.93 & 0.93 & 0.21 & 0.90 & L\\
		44 & 88.77 & 26 & 0.84 & 0.04 & 0.83 & 0.89 & 0.00 & 0.90 & L & 94 & 96.58 & 68 & 1.00 & 0.15 & 0.97 & 1.00 & 0.00 & 0.96 & ---\\ 
		45 & 88.89 & 67 & 0.82 & 0.19 & 1.00 & 1.00 & 0.04 & 0.92 & L & 95 & 96.80 & 14 & 1.00 & 0.05 & 1.00 & 0.86 & 0.06 & 0.95 & --- \\
		46 & 88.97 & 94 & 0.92 & 0.10 & 0.94 & 0.90 & 0.12 & 0.91 & L & 96 & 98.94 & 2 & 0.97 & 0.00 & 0.98 & 0.96 & 0.12 & 0.96 & ---\\ 
		47 & 89 & 63 & 0.89 & 0.04 & 0.83 & 0.93 & 0.08 & 0.91 & L & 97 & 99.05 & 24 & 1.00 & 0.07 & 1.00 & 0.92 & 0.08 & 0.95 & --- \\
		48 & 89.03 & 58 & 1.00 & 0.21 & 0.88 & 0.98 & 0.11 & 0.91 & L & 98 & 99.30 & 29 & 1.00 & 0.00 & 0.88 & 1.00 & 0.10 & 0.96 & ---\\ 
		49 & 89.13 & 46 & 0.91 & 0.23 & 0.89 & 0.93 & 0.00 & 0.90 & L & 99 & 99.56 & 87 & 0.97 & 0.12 & 0.91 & 1.00 & 0.00 & 0.95 & ---\\
		50 & 89.23 & 34 & 0.94 & 0.20 & 1.00 & 0.92 & 0.16 & 0.90 & L & 100 & 99.69 & 95 & 1.00 & 0.08 & 1.00 & 0.88 & 0.04 & 0.95 & ---\\ 		
		\hline	
	\end{tabular*}
\end{table*}
Following a successful ROI extraction, the image is binarized and dewarped. As mentioned in the previous sections, the dewarping process takes place in multiple steps. After an initial page-level dewarp, the metrics describing the parallelism, orthogonality, geodesic property, text height, as described in the previous section, are calculated. This is depicted in Table \ref{tab:step-by-step}. Based on an overall quality metric given by 
\begin{equation}
\tilde{\mu} = \dfrac{1}{5}(\overbar{\mu}_1 + (1-\overbar{\mu}_2) + \overbar{\mu}_3 + \overbar{\mu}_4 + (1-\overbar{\mu}_5)),
\label{eq:overall_metric}
\end{equation}
a decision is made on what to be done next. It is worth noting that in case of a perfectly straight document image where every single text-line is of the same height and are perfectly parallel, the following would be the ideal normalized average values of the parameters:
\begin{enumerate}
	\item orthogonality of the projection $ \mu_{1} = 1 $;
	\item parallelism of the text-lines: $ \mu_{2} = 0$;
	\item geodesic property of the lines: $ \mu_{3} = 1$;
	\item orthogonality of the text strokes and line directions: $ \mu_{4} = 1 $;
	\item metric for relative text-line height: $ \mu_{5} = 0 $.
\end{enumerate}
Based on these estimates, if it is found that the resulting image after the initial stage of page-level dewarping requires good amount of overall improvement, the page-level dewarping process is repeated once more. The metrics are calculated again. If, after one or two iterations of page-level dewarping, the quality metrics indicate that only line-level corrections are needed, then line-level dewarping is applied on the processed image. Table \ref{tab:step-by-step} is sorted by the OCR accuracy after the initial step. At each step, the quality is assessed and decisions are made in accordance with the desired quality set by the user. In the present settings, if the overall quality metric $ \tilde{\mu} $ happens to be greater than or equal to $ 0.95 $, then nothing is done after the initial application of page-level dewarping. This is indicated by ``---". If $ \tilde{\mu} $ is found to be within $ [0.90, 0.95) $, only line-level corrections are deemed necessary. This is indicated with ``L". Similarly, ``R+L" indicates that after the initial application of page-level dewarping, the process is repeated once more and then line-level corrections are applied. This is done if the value of $ \tilde{\mu} $ after the first step was found to be below $ 0.90 $. 

This is presented here just as a depiction of the correlation among the quality metrics and the OCR accuracy. In real applications, however, ASCII ground truths would obviously be unavailable and calculating OCR accuracy would be impossible. Thus, in real-world applications, these quality metrics \textit{only} can provide an estimate of the quality of the whole process in the absence of a ground truth.

\subsection{Dewarping evaluation measure}
Since the ground-truth images for the present dataset are available, the \textit{dewarping evaluation measure} ($ \mathcal{D}_M $) is also calculated as per \cite{6320850} after the completion of a full dewarping stack. Scale invariant keypoints are first marked on the text-lines of both the warped and dewarped images. Let $ p_j(x) $ and $ q_j(x) $ denote smooth cubic polynomials representing the keypoints on the $ j $-th line on a warped and dewarped image. The warp-mesures for the warped and the dewarped images, respectively, are calculated as,
\begin{equation*}
S_j=\sum_{n=1}^{k}\int_{^n x_{\textrm{start}}}^{^n x_{\textrm{end}}}p_j(x)dx, \textrm{and } \quad R_j=\sum_{n=1}^{k}\int_{^n x'_{\textrm{start}}}^{^n x'_{\textrm{end}}}q_j(x')dx.
\end{equation*}
The dewarping evaluation measure $ \mathcal{D}_{M_j} $ for the  $ j $-th line  is then defined as,
\begin{equation}\label{eq:DM}
\mathcal{D}_{M_j} = \begin{cases}
1 - \dfrac{R_j}{S_j}, & \textrm{if } \dfrac{R_j}{S_j} <1\\
0 & \textrm{otherwise}
\end{cases}.
\end{equation}
$ \mathcal{D}_{M_j} = 0 $ is assigned when the dewarped text-line is even worse than the warped one. Often it is deemed sufficient to calculate $ \mathcal{D}_{M_j}$ for a few important and critical text-lines only. The average measure, represented as $ \mathcal{D}_{M}$ over $ N $ number of lines, is defined as,
\begin{equation}\label{eq:avg-dm}
\mathcal{D}_{M} = \left( \frac{1}{N} \sum_{j=1}^{N} \mathcal{D}_{M_j} \right) \times 100\%.
\end{equation}
Also, since different text-lines would have different amounts of warps in them, a weighed average dewarping evaluation measure $ w\mathcal{D}_M $ can be calculated by modifying \eqref{eq:avg-dm} as,
\begin{equation}\label{eq:wdm}
w\mathcal{D}_M = \left(\sum_{j=1}^{N} w_j\,\mathcal{D}_{M_j} \right) \times 100\%,
\end{equation}
where, $ w_j = S_j / \sum_{l=1}^{N}S_l$, for $ \quad1\leq j\leq N  $.

\begin{table}
	\centering 
	\rowcolors{2}{white}{gray!25}
	\caption{Average dewarping evaluation measure ($ \mathcal{D}_M $)  and weighted average dewarping evaluation Measure ($ w\mathcal{D}_M $) for various algorithms on the CBDAR 2007 / IUPR 2011 document image dewarping  dataset. Higher values indicate better results.\label{tab:dm}}
	\begin{tabular*}{0.97\linewidth}{lccc}    \toprule
		\textbf{algorithm} & $ \mathcal{D}_M\,(N=6) $ & $ w\mathcal{D}_M\,(N=6) $ & $ \mathcal{D}_M $ (all) \\\midrule
		Masalovitch \cite{masalovitch2007usage} & 77.44 & 76.13 & 75.37\\
		Gatos \cite{Gatos} & 86.31 & 83.62 & 82.19 \\
		Stamatopoulos \cite{5585760} & 90.63 & 88.22 & 86.67 \\
		Wu \cite{Wu_Model} & 93.03 & 92.12 & 91.62 \\
		Bukhari \cite{Bukhari} & 93.64 & 92.71 & 92.37 \\
		Meng \cite{Meng_Metric} & 94.45 & 94.06 & 93.18 \\
		Kil \cite{Kil} & 94.71 & 93.36 & 93.08 \\
		Proposed & \textbf{96.34} & \textbf{96.11} & \textbf{95.86} \\	
		\hline
	\end{tabular*}
\end{table}
The average dewarping evaluation measure ($ \mathcal{D}_M $)  and weighted average dewarping evaluation measure ($ w\mathcal{D}_M $) for various algorithms on the CBDAR 2007 / IUPR 2011 document image dewarping  dataset is provided in Table \ref{tab:dm}. The keypoints are chosen at the beginning of each word on the text-lines for standardizing the process of evaluation over multiple set of algorithms. The table shows the average measure for 6 critical text-lines per image, the weighted average measure for 6 critical text-lines per image and the average measure for all text-lines. 

\subsection{Implementation specific details}
\begin{table}
	\centering 
	\rowcolors{2}{white}{gray!25}
	\caption{Total run-time and final OCR accuracy for each sample in the CBDAR 2007 / IUPR 2011 document image dewarping  dataset\label{tab:results}}
	\begin{tabular*}{.97\linewidth}{rccrcc}    \toprule
		\textbf{img}. & \textbf{run-time} (s) & \textbf{acc}. (\%) & \textbf{img}. & \textbf{run-time} (s) & \textbf{acc}. (\%) \\\midrule
		\textbf{1} & 7.091 & \textbf{100.00} & \textbf{51} & 7.275 & 98.29 \\
		\textbf{2} & 2.661 & 98.94 & \textbf{52} & 7.656 & 97.47 \\
		\textbf{3} & 4.049 & 98.2 & \textbf{53} & 8.262 & 97.52 \\
		\textbf{4} & 5.729 & 99.27 & \textbf{54} & 7.578 & 99.33 \\
		\textbf{5} & 6.586 & 98.17 & \textbf{55} & 7.97 & 97.81 \\
		\textbf{6} & 5.132 & 99.1 & \textbf{56} & 3.52 & 98.74 \\
		\textbf{7} & 3.873 & 97.94 & \textbf{57} & 5.667 & 98.83 \\
		\textbf{8} & 4.392 & 98.11 & \textbf{58} & 5.626 & 97.28 \\
		\textbf{9} & 4.023 & 97.21 & \textbf{59} & 5.567 & 98.2 \\
		\textbf{10} & 5.313 & 97.39 & \textbf{60} & 5.619 & 97.97 \\
		\textbf{11} & 6.377 & 99.77 & \textbf{61} & 5.759 & 99.11 \\
		\textbf{12} & 7.75 & 98.32 & \textbf{62} & 5.832 & 97.85 \\
		\textbf{13} & 7.706 & 97.06 & \textbf{63} & 6.699 & 98.27 \\
		\textbf{14} & 2.419 & 96.8 & \textbf{64} & 4.517 & 98.82 \\
		\textbf{15} & 6.855 & 98.05 & \textbf{65} & 9.135 & 96.75 \\
		\textbf{16} & 5.522 & 99.24 & \textbf{66} & 5.27 & 98.89 \\
		\textbf{17} & 5.455 & 98.75 & \textbf{67} & 5.211 & 99.41 \\
		\textbf{18} & 8.589 & 99.79 & \textbf{68} & 2.573 & 96.58 \\
		\textbf{19} & 6.646 & 98 & \textbf{69} & 3.586 & 96.73 \\
		\textbf{20} & \textbf{9.563} & 99.15 & \textbf{70} & 5.213 & 98.66 \\
		\textbf{21} & 5.892 & 98.19 & \textbf{71} & 7.285 & 98.83 \\
		\textbf{22} & 4.921 & 98.72 & \textbf{72} & 7.889 & 97.95 \\
		\textbf{23} & 5.635 & 99.69 & \textbf{73} & 7.488 & 98.65 \\
		\textbf{24} & 2.507 & 99.05 & \textbf{74} & 6.813 & 98.02 \\
		\textbf{25} & 6.154 & 99.99 & \textbf{75} & 5.593 & 98.86 \\
		\textbf{26} & 6.56 & 98.93 & \textbf{76} & 6.132 & 97.32 \\
		\textbf{27} & 8.237 & 98.61 & \textbf{77} & 7.571 & 99.22 \\
		\textbf{28} & 7.08 & 97.9 & \textbf{78} & 6.907 & 98.23 \\
		\textbf{29} & 2.276 & 99.3 & \textbf{79} & 7.227 & 98.7 \\
		\textbf{30} & 3.893 & 96.19 & \textbf{80} & 4.54 & 97.18 \\
		\textbf{31} & 4.804 & 97.55 & \textbf{81} & 6.592 & 98.9 \\
		\textbf{32} & 9.161 & 98.92 & \textbf{82} & 5.745 & 97.42 \\
		\textbf{33} & 7.749 & 98.77 & \textbf{83} & 8.85 & 98.42 \\
		\textbf{34} & 4.889 & 98.9 & \textbf{84} & 6.449 & 98.27 \\
		\textbf{35} & 6.015 & 97.75 & \textbf{85} & 5.426 & 98.28 \\
		\textbf{36} & 5.934 & 97.7 & \textbf{86} & 8.166 & 96.87 \\
		\textbf{37} & 4.728 & 95.7 & \textbf{87} & 2.633 & 99.56 \\
		\textbf{38} & 7.15 & 98.52 & \textbf{88} & 8.287 & 97.44 \\
		\textbf{39} & 5.525 & 97.03 & \textbf{89} & 8.133 & 99.85 \\
		\textbf{40} & 7.964 & 97.09 & \textbf{90} & 6.329 & 97.26 \\
		\textbf{41} & 7.875 & 98.28 & \textbf{91} & 3.23 & 98.66 \\
		\textbf{42} & 6.325 & 98.49 & \textbf{92} & 8.416 & 98.85 \\
		\textbf{43} & 4.133 & 99.77 & \textbf{93} & 5.561 & 99.83 \\
		\textbf{44} & 3.276 & 96.04 & \textbf{94} & 6.744 & 99.47 \\
		\textbf{45} & 3.655 & 98.03 & \textbf{95} & \textbf{1.567} & 99.69 \\
		\textbf{46} & 6.577 & 95.92 & \textbf{96} & 5.484 & 97.13 \\
		\textbf{47} & 5.639 & 99.23 & \textbf{97} & 7.158 & 98.58 \\
		\textbf{48} & 5.51 & 99.05 & \textbf{98} & 6.505 & 99.25 \\
		\textbf{49} & 6.121 & \textbf{95.59} & \textbf{99} & 5.46 & 97.22 \\
		\textbf{50} & 8.821 & 96.49 & \textbf{100} & 3.479 & 98.22 \\	
		\hline	
	\end{tabular*}
\end{table}
\begin{table}
	\centering 
	\rowcolors{2}{white}{gray!25}
	\caption{Average accuracy of various algorithms on the CBDAR 2007 / IUPR 2011 document image dewarping  dataset\label{tab:OCR_avg}}
	\begin{tabular*}{\linewidth}{lc@{\extracolsep{\fill}}c}    \toprule
		\textbf{algorithm} & \textbf{OCR accuracy} (\%) & \textbf{run-time} (avg$ \pm $std.) (s)\\\midrule
		Masalovitch \cite{masalovitch2007usage} & 62.47 & $ 12.33\pm8.69 $\\
		Gatos \cite{Gatos}					& 89.47 & $ 14.10\pm5.11 $\\
		Stamatopoulos \cite{5585760}		& 93.82 & $ 10.94\pm6.55 $\\
		Wu \cite{Wu_Model}					& 96.22 & $ 8.77\pm2.98 $\\
		Bukhari \cite{Bukhari}				& 96.47 & $ 14.52\pm7.76 $\\
		Meng \cite{Meng_Metric}				& 97.50 & $ 6.33\pm4.64 $\\
		Kil \cite{Kil}						& 97.54 & $ 10.22\pm2.33 $\\
		Proposed							& \textbf{98.25} & $ \textbf{5.97} \pm\textbf{1.75} $\\
		\hline
	\end{tabular*}
\end{table}
The algorithms were implemented in Python 3.7.4. The libraries used include, NumPy 1.16, SciPy 1.3, Scikit-Image 0.15, Scikit-Learn 0.21, OpenCV 4.1 and Cython 0.29. The programs were executed on a laptop with an Intel i5 6200U CPU (2.8 GHz) and 8 GB DDR3 (1600 MHz) RAM, running Ubuntu GNU/Linux 18.04. It is seen that the present methodology yields quite satisfactory results in terms speed and the quality of the OCR after dewarping. The average OCR accuracy is calculated to be 98.25\%, which is higher than the previous state-of-the-art (97.54\%) \cite{Kil}. The OCR accuracy is calculated using Tesseract 4.0.0 and comparing the results with the ASCII ground truth provided with the dataset. Table \ref{tab:results} depicts the total run-time and OCR accuracy for each individual image in the dataset. Table \ref{tab:OCR_avg} indicates the average recognition accuracy of the present algorithm in comparison to the previous ones. It is worth noting that the present methodology is reasonably faster than the previously available algorithms. The minimum, maximum, average and standard deviation of the run-time for the present implementation are 1.57 s, 9.56 s,  5.97 s and 1.75 s, respectively. The minimum and maximum run-time for the previous best implementation \cite{Kil} was 6.23 s and 15.83 s respectively.
\subsection{Tests on the DocUNet dataset}
\newcommand{\fma}{0.118\linewidth}
\newcommand{\fmaa}{0.154\linewidth}
\begin{figure*}
	\centering
	\subfloat{\label{}\includegraphics[frame, height=\fmaa]{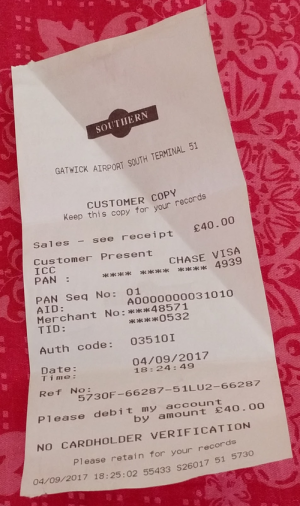}}
	\subfloat{\label{}\includegraphics[frame, height=\fmaa]{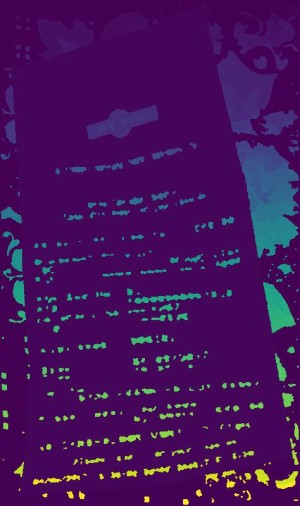}}
	\subfloat{\label{}\includegraphics[frame, height=\fmaa]{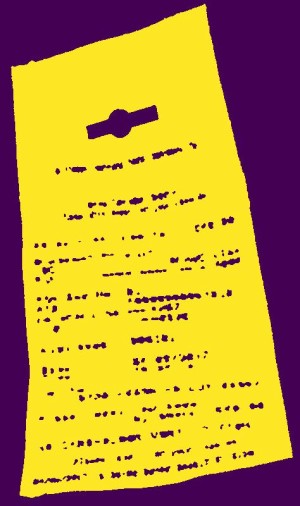}}
	\subfloat{\label{}\includegraphics[frame, height=\fmaa]{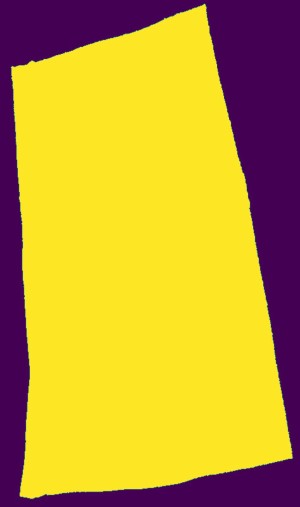}}
	\subfloat{\label{}\includegraphics[frame, height=\fmaa]{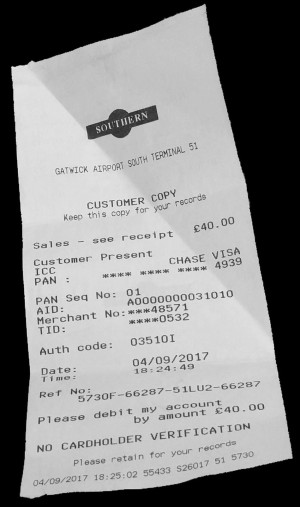}}\,
	\subfloat{\label{}\includegraphics[frame, height=\fmaa]{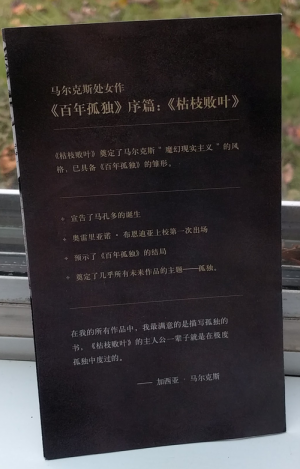}}
	\subfloat{\label{}\includegraphics[frame, height=\fmaa]{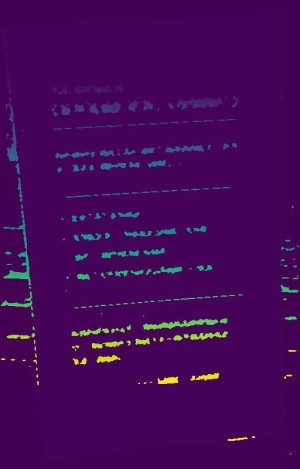}}
	\subfloat{\label{}\includegraphics[frame, height=\fmaa]{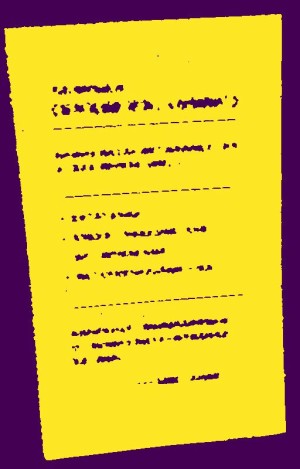}}
	\subfloat{\label{}\includegraphics[frame, height=\fmaa]{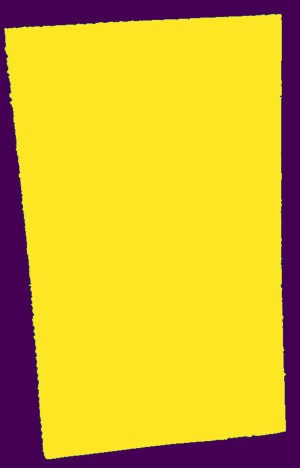}}
	\subfloat{\label{}\includegraphics[frame, height=\fmaa]{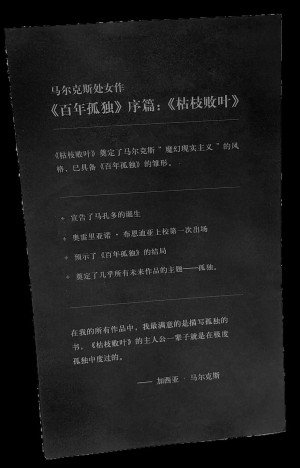}}\\\vspace{-5pt}
	\subfloat{\label{}\includegraphics[frame, height=\fma]{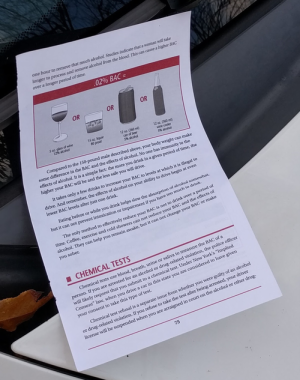}}
	\subfloat{\label{}\includegraphics[frame, height=\fma]{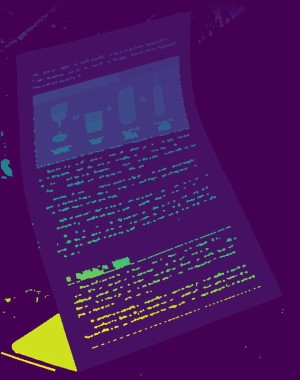}}
	\subfloat{\label{}\includegraphics[frame, height=\fma]{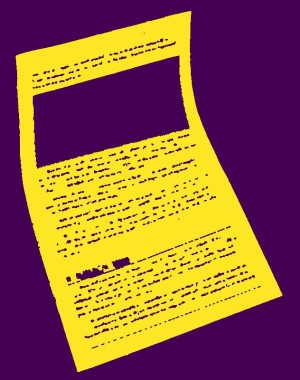}}
	\subfloat{\label{}\includegraphics[frame, height=\fma]{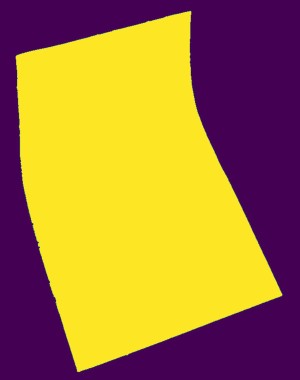}}
	\subfloat{\label{}\includegraphics[frame, height=\fma]{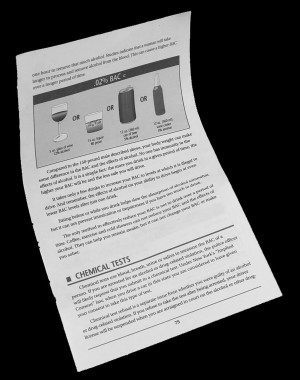}}\,
	\subfloat{\label{}\includegraphics[frame, height=\fma]{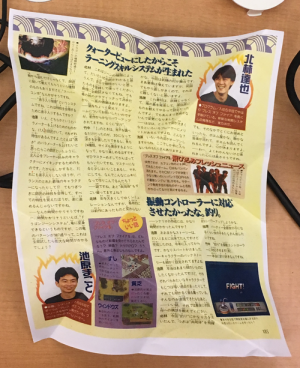}}
	\subfloat{\label{}\includegraphics[frame, height=\fma]{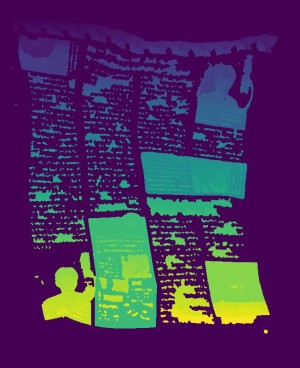}}
	\subfloat{\label{}\includegraphics[frame, height=\fma]{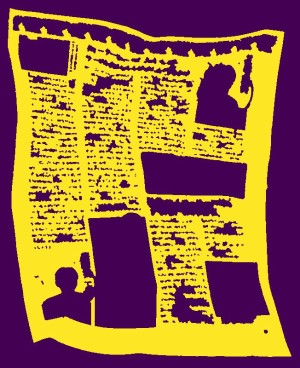}}
	\subfloat{\label{}\includegraphics[frame, height=\fma]{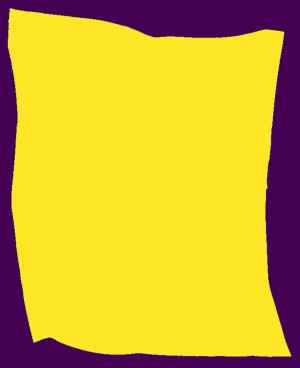}}
	\subfloat{\label{}\includegraphics[frame, height=\fma]{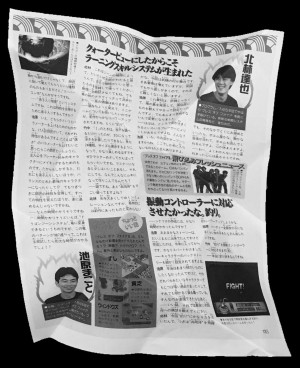}}
	\caption{\label{fig:ma_preprocessing} Sample results of ROI extraction on the DocUNet \cite{Ma_DocUNet} dataset.}
\end{figure*} 
\newcommand{\fmares}{0.15\linewidth}
\newcommand{\fmaresone}{0.18\linewidth}
\newcommand{\fmarestwo}{0.16\linewidth}
\begin{figure*}
	\centering
	\subfloat[warped]{\label{}\includegraphics[frame, height=\fmaresone]{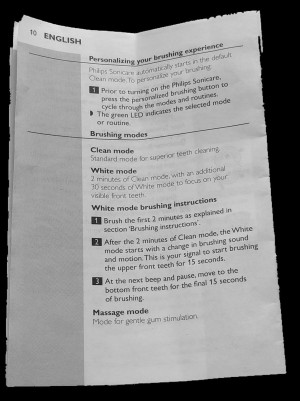}}\,
	\subfloat[dewarped]{\label{}\includegraphics[frame, height=\fmaresone]{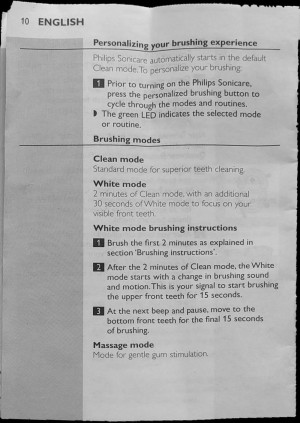}}\,
	\subfloat[warped]{\label{}\includegraphics[frame, height=\fmaresone]{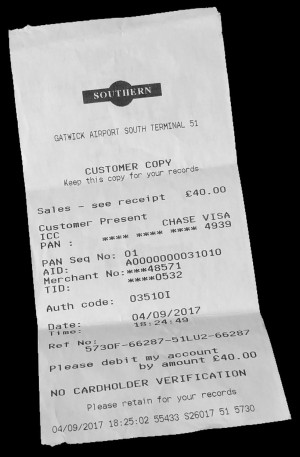}}\,
	\subfloat[dewarped]{\label{}\includegraphics[frame, height=\fmaresone]{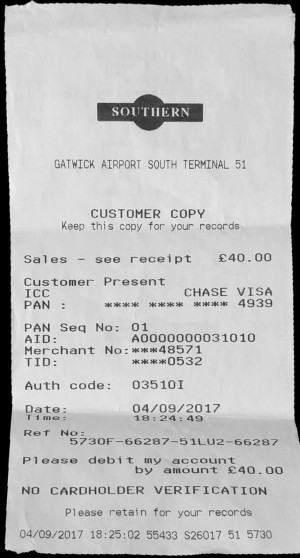}}\,
	\subfloat[warped]{\label{}\includegraphics[frame, height=\fmaresone]{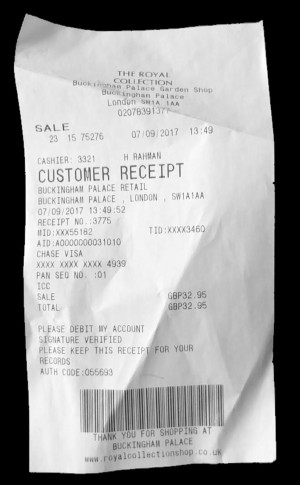}}\,
	\subfloat[dewarped]{\label{}\includegraphics[frame, height=\fmaresone]{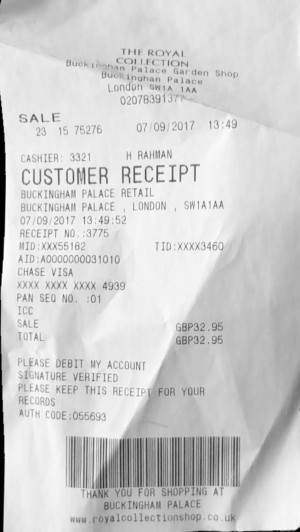}}
	\subfloat[warped]{\label{}\includegraphics[frame, height=\fmaresone]{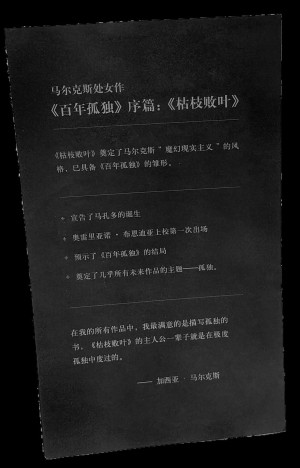}}\,
	\subfloat[dewarped]{\label{}\includegraphics[frame, height=\fmaresone]{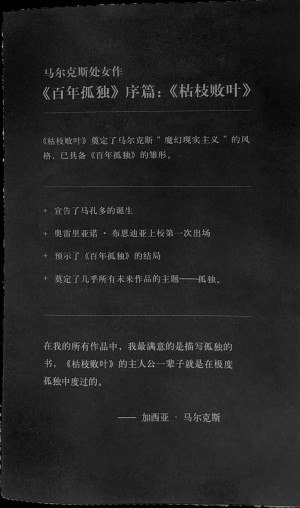}}\\
	\subfloat[warped]{\label{}\includegraphics[frame, height=\fmarestwo]{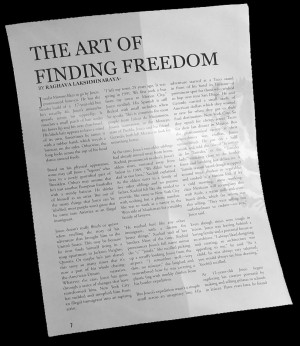}}\,
	\subfloat[dewarped]{\label{}\includegraphics[frame, height=\fmarestwo]{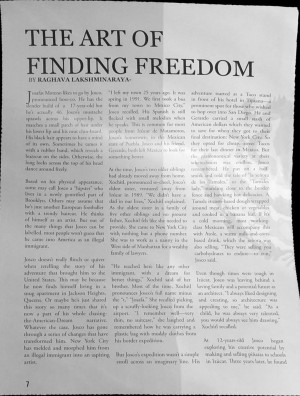}}\,
	\subfloat[warped]{\label{}\includegraphics[frame, height=\fmarestwo]{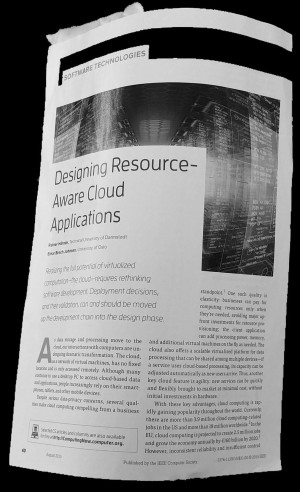}}\,
	\subfloat[dewarped]{\label{}\includegraphics[frame, height=\fmarestwo]{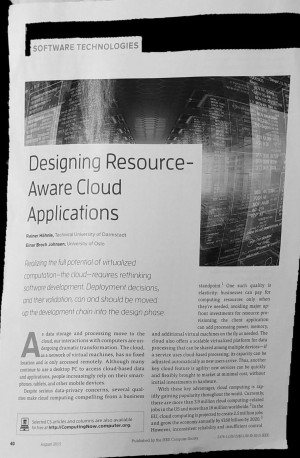}}\,
	\subfloat[warped]{\label{}\includegraphics[frame, height=\fmarestwo]{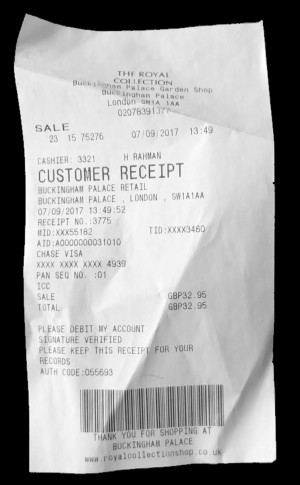}}\,
	\subfloat[dewarped]{\label{}\includegraphics[frame, height=\fmarestwo]{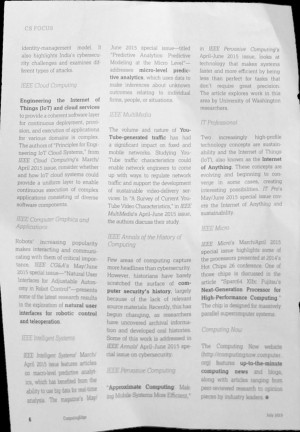}}\,
	\subfloat[warped]{\label{}\includegraphics[frame, height=\fmarestwo]{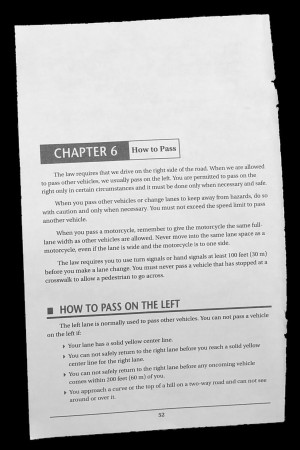}}\,
	\subfloat[dewarped]{\label{}\includegraphics[frame, height=\fmarestwo]{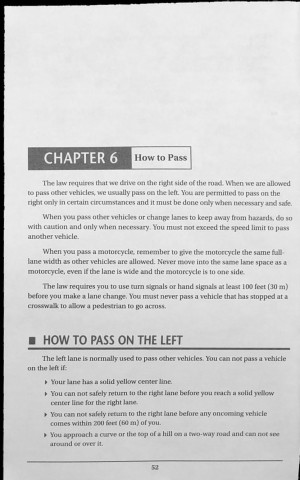}}\\
	\subfloat[warped]{\label{}\includegraphics[frame, height=\fmares]{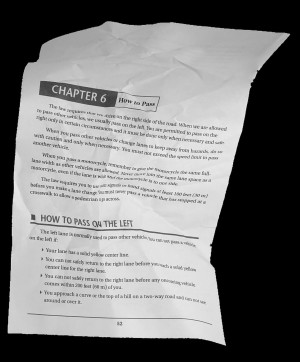}}\
	\subfloat[dewarped]{\label{}\includegraphics[frame, height=\fmares]{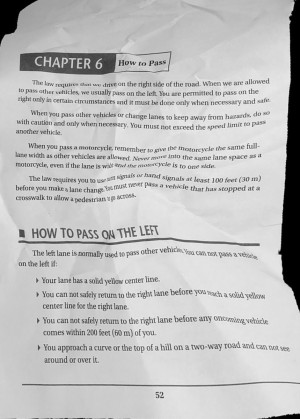}}\,
	\subfloat[warped]{\label{}\includegraphics[frame, height=\fmares]{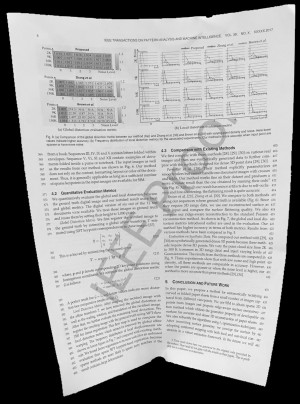}}\,
	\subfloat[dewarped]{\label{}\includegraphics[frame, height=\fmares]{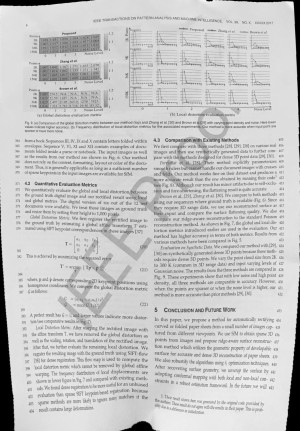}}\,
	\subfloat[warped]{\label{}\includegraphics[frame, height=\fmares]{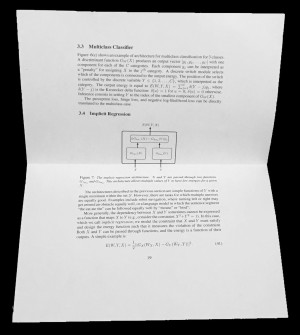}}\,
	\subfloat[dewarped]{\label{}\includegraphics[frame, height=\fmares]{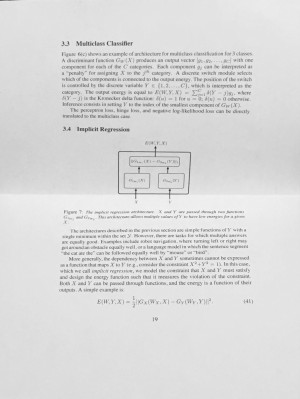}}\,
	\subfloat[warped]{\label{}\includegraphics[frame, height=\fmares]{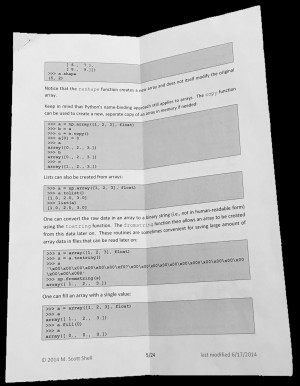}}\,
	\subfloat[dewarped]{\label{}\includegraphics[frame, height=\fmares]{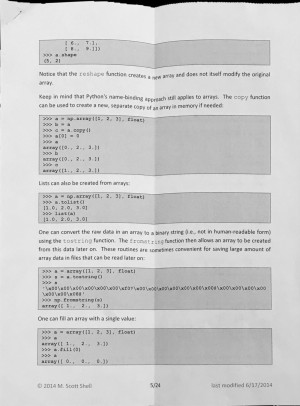}}\,
	\caption{\label{fig:ma_results} Sample results of dewarping on the images in the DocUNet \cite{Ma_DocUNet} dataset. The images are used after removing their backgrounds and converting them to grayscale.}
\end{figure*} 
As a potential validation of the effectiveness of the present algorithms, further tests are done on the DocUNet \cite{Ma_DocUNet} dataset. It contains a set of 130 synthetically warped document images representative of a multitude of (extreme) real-world scenarios. This dataset is extremely challenging for any conventional algorithm because of the inherent non-uniformity of the warps in the images. Since the present methodology partially relies on the absence of arbitrary and confusing backgrounds, the ROI detection process is deemed to be of utmost importance. However, as it can been seen from the sample results shown in fig. \ref{fig:ma_preprocessing}, the pre-processing algorithm for ROI selection does an excellent job of quickly removing the backgrounds from the images. 

Since this dataset contains warps that cannot be represented properly with only the knowledge of the page-boundaries, the present methodology is slightly tweaked. The modifications include the following assumptions:
\begin{enumerate}
	\item knowledge of the page-boundaries is mostly useless,
	\item homographies cannot be extrapolated based on the detection of just two parallel straight lines, and
	\item no smoothness criteria can be assumed in the nature of the warps.
\end{enumerate}

\begin{figure}
	\centering
	\includegraphics[width=0.9\linewidth]{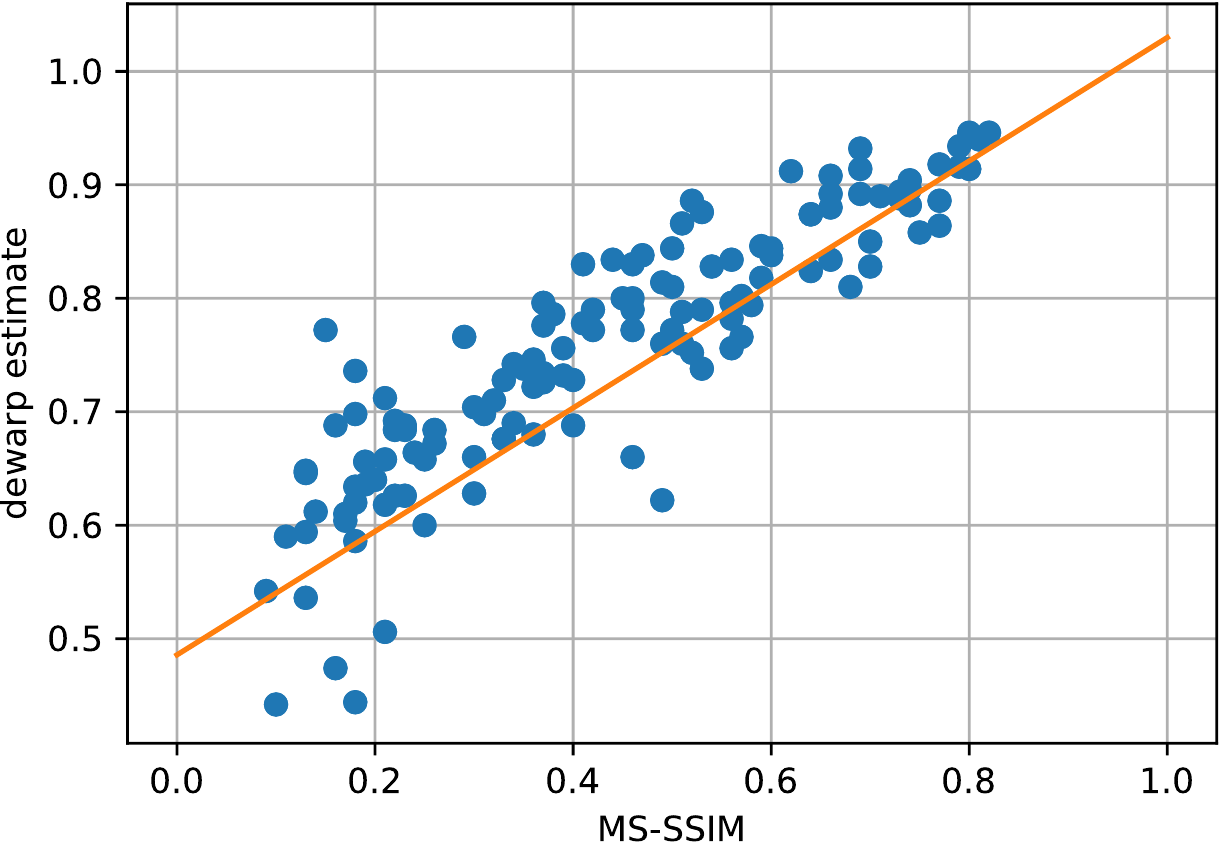}
	\caption{Plot of MS-SSIM vs dewarp estimate ($ \tilde{\mu} $) for each sample in the DocUNet 2018 dataset after one iteration of page-level dewarp. The regression line is depicted in orange.}
	\label{fig:dewarpcorr}
\end{figure}

The quality of each step in the dewarping process is again measured by calculating the orthogonality and parallelism measures. The criteria is, however, slightly relaxed in view of the inherent complexity of the images. If the overall measure as given by \eqref{eq:overall_metric}, fell under 0.8, the dewarping process was repeated with finer adjustments. If $ 0.8 \leq \tilde{\mu} < 0.9$, only line level corrections are made. Achieving $ \tilde{\mu} \geq 0.9 $ was considered a success and no further processing was deemed necessary. Sample results of the application of the present technique on the DocUNet 2018 dataset is depicted in fig. \ref{fig:ma_results}. The average dewarping evaluation measure ($ \mathcal{D}_M $)  and weighted average dewarping evaluation measure ($ w\mathcal{D}_M $) for various algorithms on the DocUNet 2018 dataset has also been calculated. This is provided in Table \ref{tab:docunet}. 
\begin{table}[h]
	\centering 
	\rowcolors{2}{white}{gray!25}
	\caption{Average dewarping evaluation measure ($ \mathcal{D}_M $)  and weighted average dewarping evaluation measure ($ w\mathcal{D}_M $) for various algorithms on the DocUNet 2018 \cite{Ma_DocUNet} dataset\label{tab:docunet}}
	\begin{tabular*}{0.97\linewidth}{lccc}    \toprule
		\textbf{algorithm} & $ \mathcal{D}_M\,(N=6) $ & $ w\mathcal{D}_M\,(N=6) $ & $ \mathcal{D}_M $ (all) \\\midrule
		Masalovitch \cite{masalovitch2007usage} & 60.39 & 58.40 & 55.21\\
		Gatos \cite{Gatos} & 63.31 & 60.20 & 58.36 \\
		Stamatopoulos \cite{5585760} & 66.60 & 64.56 & 63.21 \\
		Wu \cite{Wu_Model} & 70.24 & 68.38 & 64.28 \\
		Bukhari \cite{Bukhari} & 70.35 & 68.12 & 65.33 \\
		Meng \cite{Meng_Metric} & 72.83 & 71.93 & 69.41 \\
		Kil \cite{Kil} & 74.26 & 73.62 & 70.46 \\
		Ma \cite{Ma_DocUNet} & 79.93 & 77.54 & 76.61 \\
		Proposed & \textbf{84.69} & \textbf{80.14} & \textbf{78.60} \\	
		\hline
	\end{tabular*}
\end{table}
However, for this dataset a comparison of OCR accuracy is omitted. This has been done since the majority of the images in the DocUNet 2018 dataset are covered with figures, instead of texts. Thus, as another structural measure of dewarping accuracy, multi-scale structural similarity (MS-SSIM) \cite{Wang2003multiscale} has been calculated for the dewarped images. This provides a metric for structural similarity between the dewarped images with the ground truths. Also, scale invariant feature transform flow (SIFT flow) \cite{Liu_SIFT} has been used to measure the local distortions (LD) \cite{You_Multiview}. As a depiction of the strong correlation between dewarp estimate ($ \tilde{\mu} $) and MS-SSIM ($ \mathcal{M} $), their values are calculated for every sample in the DocUNet 2018 dataset after one iteration of page-level dewarping. The results are shown in Table \ref{tab:ma-step-by-step}. 
\begin{table*}
	\centering 
	\rowcolors{2}{white}{gray!25}
	\caption{Depiction of the strong correlation between the quality metric ($ \tilde{\mu} $) designed in this paper and the MS-SSIM ($ \mathcal{M} $) for the images in the DocUNet 2018 \cite{Ma_DocUNet} dataset. The depicted values are the calculated metrics after a single iteration of page-level dewarping. One important point to note here is that, calculating MS-SSIM requires the ground truth images, while $ \tilde{\mu} $ does not.  \label{tab:ma-step-by-step}}
	\begin{tabular*}{0.99\linewidth}{p{20pt}p{20pt}p{20pt}p{20pt}p{20pt}p{20pt}p{20pt}p{20pt}|p{20pt}p{20pt}p{20pt}p{20pt}p{20pt}p{20pt}p{20pt}p{20pt}}    \toprule
		\textbf{img.} & $ \mathcal{M} $ & $ \overbar{\mu}_1 $ &  $ \overbar{\mu}_2 $ &  $ \overbar{\mu}_3 $ &  $ \overbar{\mu}_4 $ &  $ \overbar{\mu}_5 $ & $ \tilde{\mu} $ & \textbf{img.} & $ \mathcal{M} $ & $ \overbar{\mu}_1 $ &  $ \overbar{\mu}_2 $ &  $ \overbar{\mu}_3 $ &  $ \overbar{\mu}_4 $ &  $ \overbar{\mu}_5 $ & $ \tilde{\mu} $\\\midrule
		1\_1 & 0.60 & 0.84 & 0.27 & 0.85 & 0.84 & 0.07 & 0.84 & 33\_2 & 0.51 & 0.57 & 0.10 & 0.68 & 0.99 & 0.20 & 0.79\\ 
		1\_2 & 0.22 & 0.24 & 0.03 & 0.66 & 0.63 & 0.37 & 0.63 & 34\_1 & 0.40 & 0.59 & 0.10 & 0.50 & 0.87 & 0.22 & 0.73\\ 
		2\_1 & 0.16 & 0.81 & 0.89 & 0.26 & 0.63 & 0.44 & 0.47 & 34\_2 & 0.69 & 0.98 & 0.13 & 0.96 & 0.87 & 0.02 & 0.93\\ 
		2\_2 & 0.23 & 0.64 & 0.09 & 0.43 & 0.47 & 0.02 & 0.69 & 35\_1 & 0.50 & 0.83 & 0.09 & 0.79 & 0.76 & 0.24 & 0.81\\ 
		3\_1 & 0.52 & 0.98 & 0.18 & 0.92 & 0.90 & 0.19 & 0.89 & 35\_2 & 0.66 & 0.86 & 0.12 & 0.94 & 0.99 & 0.21 & 0.89\\ 
		3\_2 & 0.47 & 0.65 & 0.09 & 0.96 & 0.71 & 0.04 & 0.84 & 36\_1 & 0.09 & 0.35 & 0.32 & 0.44 & 0.59 & 0.35 & 0.54\\ 
		4\_1 & 0.74 & 0.77 & 0.14 & 0.92 & 0.87 & 0.01 & 0.88 & 36\_2 & 0.13 & 0.24 & 0.47 & 0.55 & 1.00 & 0.35 & 0.59\\ 
		4\_2 & 0.13 & 0.27 & 0.02 & 0.94 & 0.60 & 0.55 & 0.65 & 37\_1 & 0.46 & 0.79 & 0.17 & 0.83 & 0.78 & 0.08 & 0.83\\ 
		5\_1 & 0.56 & 0.84 & 0.19 & 0.61 & 0.68 & 0.03 & 0.78 & 37\_2 & 0.77 & 0.91 & 0.13 & 0.87 & 0.90 & 0.12 & 0.89\\ 
		5\_2 & 0.37 & 0.42 & 0.23 & 0.92 & 0.85 & 0.29 & 0.73 & 38\_1 & 0.18 & 0.40 & 0.19 & 0.69 & 0.83 & 0.24 & 0.70\\ 
		6\_1 & 0.54 & 0.84 & 0.25 & 0.71 & 0.85 & 0.01 & 0.83 & 38\_2 & 0.39 & 0.83 & 0.43 & 0.48 & 0.98 & 0.08 & 0.76\\ 
		6\_2 & 0.49 & 0.52 & 0.29 & 0.59 & 0.63 & 0.34 & 0.62 & 39\_1 & 0.73 & 0.79 & 0.14 & 0.98 & 0.98 & 0.14 & 0.89\\ 
		7\_1 & 0.82 & 0.94 & 0.01 & 0.93 & 0.98 & 0.11 & 0.95 & 39\_2 & 0.35 & 0.80 & 0.37 & 0.74 & 0.92 & 0.40 & 0.74\\ 
		7\_2 & 0.66 & 0.74 & 0.05 & 0.70 & 0.84 & 0.06 & 0.83 & 40\_1 & 0.30 & 0.35 & 0.36 & 0.96 & 0.45 & 0.10 & 0.66\\ 
		8\_1 & 0.75 & 0.85 & 0.17 & 0.90 & 0.86 & 0.15 & 0.86 & 40\_2 & 0.51 & 1.00 & 0.08 & 0.78 & 0.81 & 0.18 & 0.87\\ 
		8\_2 & 0.68 & 0.74 & 0.14 & 0.81 & 0.78 & 0.14 & 0.81 & 41\_1 & 0.17 & 0.69 & 0.26 & 0.55 & 0.41 & 0.34 & 0.61\\ 
		9\_1 & 0.77 & 0.93 & 0.03 & 0.82 & 0.88 & 0.01 & 0.92 & 41\_2 & 0.60 & 0.90 & 0.19 & 0.87 & 0.84 & 0.20 & 0.84\\ 
		9\_2 & 0.80 & 0.93 & 0.06 & 0.86 & 0.85 & 0.01 & 0.91 & 42\_1 & 0.42 & 0.46 & 0.02 & 0.87 & 0.96 & 0.32 & 0.79\\ 
		10\_1 & 0.69 & 0.95 & 0.18 & 0.89 & 1.00 & 0.20 & 0.89 & 42\_2 & 0.80 & 0.96 & 0.07 & 0.97 & 0.89 & 0.02 & 0.95\\ 
		10\_2 & 0.56 & 0.86 & 0.12 & 0.85 & 0.71 & 0.13 & 0.83 & 43\_1 & 0.33 & 0.95 & 0.06 & 0.52 & 0.57 & 0.34 & 0.73\\ 
		11\_1 & 0.37 & 0.59 & 0.08 & 0.82 & 0.64 & 0.34 & 0.73 & 43\_2 & 0.18 & 0.80 & 0.33 & 0.39 & 0.54 & 0.47 & 0.59\\ 
		11\_2 & 0.36 & 0.52 & 0.20 & 0.50 & 0.77 & 0.19 & 0.68 & 44\_1 & 0.46 & 0.99 & 0.16 & 0.54 & 0.86 & 0.23 & 0.80\\ 
		12\_1 & 0.39 & 0.72 & 0.25 & 0.66 & 0.73 & 0.20 & 0.73 & 44\_2 & 0.30 & 0.48 & 0.47 & 0.69 & 0.92 & 0.10 & 0.70\\ 
		12\_2 & 0.19 & 0.66 & 0.34 & 0.43 & 0.69 & 0.16 & 0.66 & 45\_1 & 0.81 & 0.95 & 0.03 & 0.84 & 0.95 & 0.01 & 0.94\\ 
		13\_1 & 0.70 & 0.86 & 0.08 & 0.76 & 0.86 & 0.15 & 0.85 & 45\_2 & 0.59 & 0.81 & 0.27 & 0.89 & 0.92 & 0.26 & 0.82\\ 
		13\_2 & 0.58 & 0.58 & 0.17 & 0.98 & 0.84 & 0.26 & 0.79 & 46\_1 & 0.16 & 0.75 & 0.13 & 0.75 & 0.66 & 0.59 & 0.69\\ 
		14\_1 & 0.32 & 0.88 & 0.29 & 0.67 & 0.57 & 0.28 & 0.71 & 46\_2 & 0.46 & 0.48 & 0.22 & 0.78 & 0.94 & 0.12 & 0.77\\ 
		14\_2 & 0.33 & 0.41 & 0.12 & 0.51 & 0.99 & 0.41 & 0.68 & 47\_1 & 0.71 & 0.84 & 0.11 & 0.82 & 0.91 & 0.01 & 0.89\\ 
		15\_1 & 0.73 & 0.76 & 0.07 & 0.84 & 0.94 & 0.03 & 0.89 & 47\_2 & 0.33 & 0.61 & 0.33 & 0.82 & 0.67 & 0.39 & 0.68\\ 
		15\_2 & 0.24 & 0.52 & 0.35 & 0.72 & 0.79 & 0.36 & 0.66 & 48\_1 & 0.57 & 0.83 & 0.12 & 0.66 & 0.69 & 0.23 & 0.77\\ 
		16\_1 & 0.79 & 0.92 & 0.00 & 0.82 & 0.84 & 0.00 & 0.92 & 48\_2 & 0.64 & 0.89 & 0.13 & 0.88 & 0.84 & 0.11 & 0.87\\ 
		16\_2 & 0.40 & 0.87 & 0.33 & 0.64 & 0.67 & 0.41 & 0.69 & 49\_1 & 0.15 & 0.83 & 0.21 & 0.90 & 0.46 & 0.12 & 0.77\\ 
		17\_1 & 0.50 & 0.95 & 0.16 & 0.83 & 0.92 & 0.32 & 0.84 & 49\_2 & 0.18 & 0.77 & 0.26 & 0.56 & 0.37 & 0.34 & 0.62\\ 
		17\_2 & 0.14 & 0.58 & 0.14 & 0.80 & 0.42 & 0.60 & 0.61 & 50\_1 & 0.42 & 0.77 & 0.24 & 0.56 & 0.81 & 0.04 & 0.77\\ 
		18\_1 & 0.49 & 0.92 & 0.08 & 0.58 & 0.81 & 0.16 & 0.81 & 50\_2 & 0.11 & 0.55 & 0.36 & 0.71 & 0.42 & 0.37 & 0.59\\ 
		18\_2 & 0.64 & 0.81 & 0.12 & 0.81 & 0.92 & 0.05 & 0.87 & 51\_1 & 0.41 & 0.97 & 0.10 & 0.82 & 0.80 & 0.34 & 0.83\\ 
		19\_1 & 0.52 & 0.77 & 0.25 & 0.79 & 0.63 & 0.18 & 0.75 & 51\_2 & 0.37 & 0.88 & 0.22 & 0.73 & 0.75 & 0.16 & 0.80\\ 
		19\_2 & 0.53 & 0.99 & 0.26 & 0.63 & 0.69 & 0.10 & 0.79 & 52\_1 & 0.50 & 0.76 & 0.21 & 0.89 & 0.66 & 0.24 & 0.77\\ 
		20\_1 & 0.57 & 0.74 & 0.08 & 0.68 & 0.86 & 0.19 & 0.80 & 52\_2 & 0.18 & 0.14 & 0.42 & 0.56 & 0.35 & 0.41 & 0.44\\ 
		20\_2 & 0.10 & 0.31 & 0.52 & 0.51 & 0.44 & 0.53 & 0.44 & 53\_1 & 0.79 & 0.96 & 0.06 & 0.82 & 0.97 & 0.02 & 0.93\\ 
		21\_1 & 0.56 & 0.89 & 0.24 & 0.72 & 0.74 & 0.13 & 0.80 & 53\_2 & 0.69 & 0.80 & 0.15 & 0.97 & 0.99 & 0.04 & 0.91\\ 
		21\_2 & 0.29 & 0.48 & 0.46 & 1.00 & 0.81 & 0.00 & 0.77 & 54\_1 & 0.53 & 0.54 & 0.11 & 0.67 & 0.80 & 0.21 & 0.74\\ 
		22\_1 & 0.46 & 0.52 & 0.20 & 0.56 & 0.70 & 0.28 & 0.66 & 54\_2 & 0.23 & 0.99 & 0.53 & 0.30 & 0.71 & 0.05 & 0.68\\ 
		22\_2 & 0.51 & 0.59 & 0.17 & 0.67 & 0.93 & 0.22 & 0.76 & 55\_1 & 0.25 & 0.54 & 0.34 & 0.86 & 0.66 & 0.43 & 0.66\\ 
		23\_1 & 0.66 & 0.80 & 0.10 & 0.99 & 0.80 & 0.09 & 0.88 & 55\_2 & 0.20 & 0.27 & 0.19 & 0.43 & 0.83 & 0.14 & 0.64\\ 
		23\_2 & 0.19 & 0.43 & 0.24 & 0.58 & 0.54 & 0.13 & 0.64 & 56\_1 & 0.36 & 0.88 & 0.32 & 0.64 & 0.68 & 0.15 & 0.75\\ 
		24\_1 & 0.46 & 0.77 & 0.31 & 0.90 & 0.89 & 0.30 & 0.79 & 56\_2 & 0.30 & 0.79 & 0.44 & 0.55 & 0.45 & 0.21 & 0.63\\ 
		24\_2 & 0.18 & 0.76 & 0.49 & 0.69 & 0.77 & 0.05 & 0.74 & 57\_1 & 0.74 & 0.97 & 0.11 & 0.93 & 0.83 & 0.14 & 0.90\\ 
		25\_1 & 0.59 & 0.78 & 0.04 & 0.89 & 0.78 & 0.18 & 0.85 & 57\_2 & 0.70 & 0.73 & 0.14 & 0.82 & 0.82 & 0.09 & 0.83\\ 
		25\_2 & 0.64 & 0.66 & 0.14 & 0.95 & 0.72 & 0.07 & 0.82 & 58\_1 & 0.62 & 0.93 & 0.09 & 0.86 & 0.88 & 0.02 & 0.91\\ 
		26\_1 & 0.21 & 0.23 & 0.33 & 0.72 & 0.83 & 0.16 & 0.66 & 58\_2 & 0.41 & 0.64 & 0.22 & 0.85 & 0.97 & 0.35 & 0.78\\ 
		26\_2 & 0.23 & 0.57 & 0.37 & 0.73 & 0.75 & 0.24 & 0.69 & 59\_1 & 0.56 & 0.57 & 0.27 & 0.71 & 0.79 & 0.02 & 0.76\\ 
		27\_1 & 0.45 & 0.50 & 0.07 & 0.96 & 0.99 & 0.38 & 0.80 & 59\_2 & 0.21 & 0.14 & 0.17 & 0.20 & 0.39 & 0.03 & 0.51\\ 
		27\_2 & 0.36 & 0.80 & 0.19 & 0.67 & 0.64 & 0.31 & 0.72 & 60\_1 & 0.18 & 0.26 & 0.12 & 0.48 & 0.78 & 0.23 & 0.63\\ 
		28\_1 & 0.37 & 0.51 & 0.22 & 0.63 & 0.97 & 0.01 & 0.78 & 60\_2 & 0.17 & 0.55 & 0.46 & 0.62 & 0.44 & 0.13 & 0.60\\ 
		28\_2 & 0.13 & 0.28 & 0.07 & 0.49 & 0.39 & 0.41 & 0.54 & 61\_1 & 0.23 & 0.58 & 0.46 & 0.67 & 0.56 & 0.22 & 0.63\\ 
		29\_1 & 0.21 & 0.87 & 0.10 & 0.32 & 0.76 & 0.29 & 0.71 & 61\_2 & 0.26 & 0.59 & 0.04 & 0.47 & 0.81 & 0.47 & 0.67\\ 
		29\_2 & 0.26 & 0.62 & 0.36 & 0.71 & 0.68 & 0.23 & 0.68 & 62\_1 & 0.22 & 0.57 & 0.08 & 0.60 & 0.89 & 0.52 & 0.69\\ 
		30\_1 & 0.53 & 0.93 & 0.08 & 0.90 & 0.69 & 0.06 & 0.88 & 62\_2 & 0.21 & 0.60 & 0.51 & 0.75 & 0.48 & 0.23 & 0.62\\ 
		30\_2 & 0.66 & 0.96 & 0.04 & 0.89 & 0.83 & 0.10 & 0.91 & 63\_1 & 0.44 & 0.99 & 0.15 & 0.83 & 0.74 & 0.24 & 0.83\\ 
		31\_1 & 0.13 & 0.43 & 0.15 & 0.55 & 0.70 & 0.30 & 0.65 & 63\_2 & 0.25 & 0.26 & 0.28 & 0.90 & 0.40 & 0.28 & 0.60\\ 
		31\_2 & 0.31 & 0.63 & 0.44 & 0.72 & 0.64 & 0.06 & 0.70 & 64\_1 & 0.74 & 0.88 & 0.03 & 0.82 & 0.93 & 0.08 & 0.90\\ 
		32\_1 & 0.34 & 0.96 & 0.20 & 0.70 & 0.51 & 0.26 & 0.74 & 64\_2 & 0.38 & 0.74 & 0.19 & 0.71 & 0.91 & 0.24 & 0.79\\ 
		32\_2 & 0.22 & 0.83 & 0.50 & 0.65 & 0.58 & 0.14 & 0.68 & 65\_1 & 0.77 & 0.91 & 0.15 & 0.83 & 0.84 & 0.11 & 0.86\\ 
		33\_1 & 0.34 & 0.84 & 0.04 & 0.41 & 0.60 & 0.36 & 0.69 & 65\_2 & 0.49 & 0.84 & 0.34 & 0.64 & 0.71 & 0.05 & 0.76\\  		
		\hline	
	\end{tabular*}
\end{table*}
This correlation is further depicted in fig. \ref{fig:dewarpcorr}. The ordinary least squares regression line is calculated to be $\tilde{\mu} = 0.54\mathcal{M} + 0.49$.

The final average MS-SSIM and LD are provided in Table \ref{tab:ma_ssim}. Compared to the benchmark results \cite{Ma_DocUNet}, the average MS-SSIM improved from 0.41 to 0.48 and the average LD improved from 14.08 to 12.28. This particularly demonstrates the robustness of the present technique. The run-time statistics are provided in Table \ref{tab:ma_avg}.
\begin{table}[h]
	\centering 
	\rowcolors{2}{white}{gray!25}
	\caption{Multi-scale structural similarity (MS-SSIM) and local distortion (LD) of various dewarping techniques on the DocUNet 2018 \cite{Ma_DocUNet} dataset. Higher MS-SSIM indicates higher similarity with the ground truth. Lower LD indicates lower distortion compared to the ground truth.\label{tab:ma_ssim}}
	\begin{tabular*}{\linewidth}{p{90pt}p{65pt}p{62pt}}    \toprule
		\textbf{algorithm} & \textbf{MS-SSIM} & \textbf{LD} \\\midrule
		Masalovitch \cite{masalovitch2007usage}  & $ 0.24$  & $28.45 $\\
		Gatos \cite{Gatos}					& $ 0.26$  & $24.66 $\\
		Stamatopoulos \cite{5585760}		& $ 0.29$  & $21.72 $\\
		Wu \cite{Wu_Model}					& $ 0.33$  & $19.02 $\\
		Bukhari \cite{Bukhari}				& $ 0.34$  & $19.36 $\\
		Meng \cite{Meng_Metric}				& $ 0.36$  & $15.70 $\\
		Kil \cite{Kil}						& $ 0.37$  & $14.91 $\\
		Ma \cite{Ma_DocUNet} 				& $ 0.41$   & $ 14.08 $\\
		Proposed							& $ \textbf{0.48}$ & $\textbf{12.28} $\\
		\hline
	\end{tabular*}
\end{table}
\begin{table}[h]
	\centering 
	\rowcolors{2}{white}{gray!25}
	\caption{Run-time statistics for the DocUNet 2018 \cite{Ma_DocUNet} dataset\label{tab:ma_avg}}
	\begin{tabular*}{\linewidth}{lr@{\extracolsep{\fill}}r}    \toprule
		\textbf{algorithm} & \textbf{run-time average} (s) & \textbf{run-time std. dev.} (s) \\\midrule
		Masalovitch \cite{masalovitch2007usage}  & $ 9.24$  & $5.58 $\\
		Gatos \cite{Gatos}					& $ 26.78$  & $13.25 $\\
		Stamatopoulos \cite{5585760}		& $ 10.94$  & $6.55 $\\
		Wu \cite{Wu_Model}					& $ 12.65$  & $8.94 $\\
		Bukhari \cite{Bukhari}				& $ 22.70$  & $8.66 $\\
		Meng \cite{Meng_Metric}				& $ 24.60$  & $10.29 $\\
		Kil \cite{Kil}						& $ 20.67$  & $2.33 $\\
		Ma \cite{Ma_DocUNet} 				& $ 8.36$   & $ 4.68 $\\
		Proposed							& $ \textbf{5.35}$ & $\textbf{3.22} $\\
		\hline
	\end{tabular*}
\end{table}

\section{Conclusive Remarks}
As it can be seen, the proposed quality measure based on orthogonality and parallelism does an excellent job of providing an estimate for the quality of the dewarping process. The most important fact is that, this measure is not reliant upon the existence of a ground-truth of any sort. Even though the methodology proposed in this paper is designed by carefully analyzing the CBDAR 2007 / IUPR 2011 dataset, the inherent robustness of the technique still allows it to be applicable on a completely different dataset (DocUNet 2018) with a vastly different set of warps and obtained comparable results. This is despite the fact that the benchmark implementation \cite{Ma_DocUNet} clearly assumes that conventional handcrafted algorithms, that do not employ deep learning cannot produce better results. 

With that said, there are still huge developmental possibilities in this domain through the application of deep learning. But, one of the foremost challenges in this regard is the lack of suitable training datasets. One approach would be to capture a few thousand real-world samples of warped document images under different conditions and use those to generate a large number of synthetically warped document images through a generative adversarial network. 

\ifCLASSOPTIONcaptionsoff
\newpage
\fi

\balance

\bibliographystyle{IEEEtran}
\bibliography{main}

\begin{IEEEbiography}[{\includegraphics[width=1in,height=1.25in,clip,keepaspectratio]{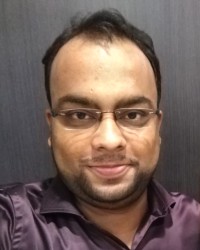}}]{Tanmoy Dasgupta}
	(S'16, M'20) received his B.E. and M.E. in electrical engineering from Jadavpur University in 2012 and 2014, respectively. He was the recipient of university gold medal in 2014 for securing first position in M.E. He is currently working as an assistant professor at Techno India University, West Bengal, India. He is presently enrolled as a doctoral student in the Department of Computer Science and Engineering, Jadavpur University. His research interests include signal and image processing.
\end{IEEEbiography}	

\vskip -2\baselineskip plus -1fil

\begin{IEEEbiography}[{\includegraphics[width=1in,height=1.25in,clip,keepaspectratio]{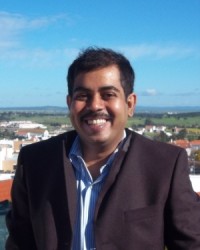}}]{Nibaran Das}
	(M'07) received his B.Tech degree in computer science and technology from Kalyani Govt. Engineering College under Kalyani University, in 2003. He received his M.C.S.E and Ph.D. (Engg.) degree from Jadavpur University, in 2005 and 2012
	respectively. He joined Jadavpur University as a faculty member in 2006 where he is currently serving as an associate professor in the department of computer science and engineering. He has published more than 150 peer-reviewed research articles in different international journals and conferences. His areas of current research interest are OCR of handwritten text, optimization techniques, deep learning and image processing. 
\end{IEEEbiography}

\vskip -2\baselineskip plus -1fil

\begin{IEEEbiography}[{\includegraphics[width=1in,height=1.25in,clip,keepaspectratio]{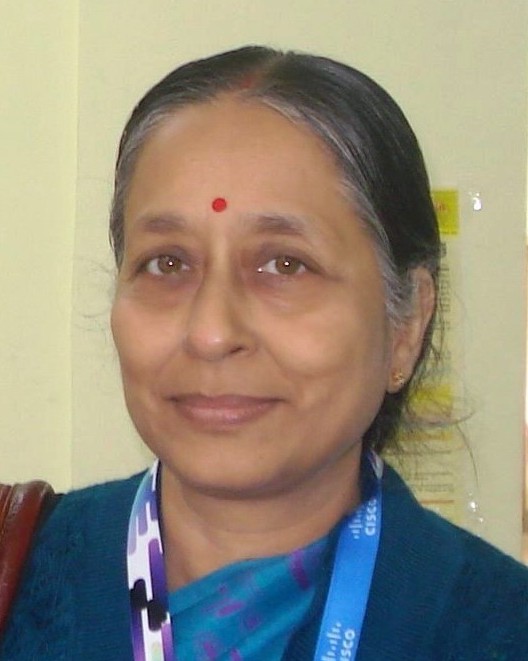}}]{Mita Nasipuri}
	(M'88, SM'92) received her B.E.Tel.E., M.E.Tel.E., and Ph.D. (Engg.) degrees from Jadavpur University, in 1979, 1981 and 1990, respectively. Prof. Nasipuri has been a faculty member of Jadavpur University since 1987. She has supervised 18 doctoral students and has published 5 books and more than 400 peer-reviewed research articles in different international journals and conferences. Her current research interest includes image processing, pattern recognition, and multimedia systems. She is a senior member of the IEEE, fellow of IE (India) and WBAST, Kolkata, India.
\end{IEEEbiography}

\end{document}